\def\paperID{0000}
\def\confName{ICCV}
\def\confYear{2025}

\def\paperTitle{
    \raisebox{-0.25\height}{\includegraphics[height=1.5em]{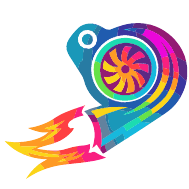}}~
    TurboReg: TurboClique for Robust and Efficient Point Cloud Registration}

\def\authorBlock{
    Shaocheng Yan\textsuperscript{1} \qquad
    Pengcheng Shi\textsuperscript{1}\footnotemark[2] \qquad 
    Zhenjun Zhao\textsuperscript{2} \\
    Kaixin Wang\textsuperscript{3} \qquad
    Kuang Cao\textsuperscript{1} \qquad
    Ji Wu\textsuperscript{4} \qquad
    Jiayuan Li\textsuperscript{1}\footnotemark[2] \\
    \textsuperscript{1}School of Remote Sensing and Information Engineering, Wuhan University \\
    \textsuperscript{2}Department of Computer and Systems Engineering, University of Zaragoza \\
    \textsuperscript{3}College of Computer Science, Beijing University of Technology \\
    \textsuperscript{4}School of Computer Science, Wuhan University \\
    {\tt\small \{shaochengyan, shipc, ljy\}@whu.edu.cn}, \\
}

\newif\ifreview 
\newif\ifarxiv \newcommand{\arxiv}{\arxivtrue}
\newif\ifcamera 
\newif\ifrebuttal 

\arxiv

\pdfoutput=1
\documentclass[10pt,twocolumn,letterpaper]{article}
\ifreview \usepackage[review]{cvpr} \fi
\ifarxiv \usepackage[pagenumbers]{cvpr} \fi
\ifrebuttal \usepackage[rebuttal]{cvpr} \fi
\ifcamera \usepackage{cvpr} \fi

\usepackage{graphicx}	
\usepackage{amsmath}	
\usepackage{amssymb}	
\usepackage{booktabs}
\usepackage{times}
\usepackage{microtype}
\usepackage{epsfig}
\usepackage{caption}
\usepackage{float}
\usepackage{placeins}
\usepackage{color, colortbl}
\usepackage{stfloats}
\usepackage{enumitem}
\usepackage{tabularx}
\usepackage{xstring}
\usepackage{multirow}
\usepackage{xspace}
\usepackage{url}
\usepackage{subcaption}
\usepackage{xcolor}
\usepackage[hang,flushmargin]{footmisc}

\ifcamera \usepackage[accsupp]{axessibility} \fi

% ...

\ifarxiv  \fi

\newcommand{\R}[1]{{%
    \textbf{%
        \ifstrequal{#1}{1}{\textcolor{red}{R#1}}{%
        \ifstrequal{#1}{2}{\textcolor{blue}{R#1}}{%
        \ifstrequal{#1}{3}{\textcolor{magenta}{R#1}}{%
        \ifstrequal{#1}{4}{\textcolor{teal}{R#1}}{%
                           \textcolor{cyan}{R#1}%
        }}}}%
    }%
}}

\usepackage{xr-hyper}

\makeatletter
\newcommand*{\addFileDependency}[1]{
  \typeout{(#1)}
  \@addtofilelist{#1}
  \IfFileExists{#1}{}{\typeout{No file #1.}}
}

\makeatother
\newcommand*{\myexternaldocument}[1]{
    \externaldocument{#1}
    \addFileDependency{#1.tex}
    \addFileDependency{#1.aux}
}

\definecolor{cvprblue}{rgb}{0.21,0.49,0.74}
\usepackage[pagebackref,breaklinks,colorlinks,allcolors=cvprblue]{hyperref}
\usepackage[capitalize]{cleveref}
\crefname{section}{Sec.}{Secs.}
\crefname{table}{Table}{Tables}
\crefname{figure}{Fig.}{Figs.}

\ifarxiv \crefname{appendix}{App.}{Apps.}
\else \crefname{appendix}{Suppl.}{Suppls.} \fi

\unless\ifarxiv \myexternaldocument{_supplementary} \fi

\usepackage{amssymb}
\usepackage{amsmath}
\usepackage[linesnumbered,ruled,vlined]{algorithm2e}
\usepackage{makecell}
\usepackage{enumitem}
\usepackage[accsupp]{axessibility}

\newtheorem{definition}{Definition}
\usepackage[dvipsnames]{xcolor}
\definecolor{mygreen}{HTML}{8E6182} 
\definecolor{myorange}{HTML}{B64645} 

\newcommand{\PAR}[1]{\vskip4pt \noindent{\bf #1}}

\newcommand{\cyan}[1]{\textcolor{cyan}{#1}}
\newcommand{\red}[1]{\textcolor{red}{#1}}

\newcommand{\redll}[1]{\textcolor{red}{#1}}
\newcommand{\redx}[1]{\textcolor{red}{#1}}

\renewcommand{\cyan}[1]{#1}
\renewcommand{\red}[1]{#1}

\renewcommand{\redx}[1]{#1}
\renewcommand{\redll}[1]{#1}

\newcommand{\annot}[1]{\textcolor{gray!80}{\small \% #1}}

\begin{document}
\title{\paperTitle}
\author{\authorBlock}
\maketitle

\makeatletter
\def\blfootnote{\xdef\@thefnmark{}\@footnotetext}
\makeatother
\blfootnote{$^\dag$ indicates the corresponding author (Jiayuan Li and Pengcheng Shi).}

\begin{abstract}
Robust estimation is essential in correspondence-based Point Cloud Registration (PCR). Existing methods using maximal clique search in compatibility graphs achieve high recall but suffer from exponential time complexity, limiting their use in time-sensitive applications. To address this challenge, we propose a fast and robust estimator, TurboReg, built upon a novel lightweight clique, TurboClique, and a highly parallelizable Pivot-Guided Search (PGS) algorithm. First, we define the TurboClique as a 3-clique within a highly-constrained compatibility graph. The lightweight nature of the 3-clique allows for efficient parallel searching, and the highly-constrained compatibility graph ensures robust spatial consistency for stable transformation estimation. Next, PGS selects matching pairs with high SC$^2$ scores as pivots, effectively guiding the search toward TurboCliques with higher inlier ratios. Moreover, the PGS algorithm has linear time complexity and is significantly more efficient than the maximal clique search with exponential time complexity. Extensive experiments show that TurboReg achieves state-of-the-art performance across multiple real-world datasets, with substantial speed improvements. 
For example, on the 3DMatch+FCGF dataset, TurboReg (1K) operates $208.22\times$ faster than 3DMAC while also achieving higher recall. Our code is accessible at \href{https://github.com/Laka-3DV/TurboReg}{\texttt{TurboReg}}.
\end{abstract}
\section{Introduction}
\label{sec:intro}

Point cloud registration (PCR) aims to align 3D scans from different viewpoints of the same scene, which is essential for tasks like simultaneous localization and mapping (SLAM)~\cite{bailey2006simultaneous, durrant2006simultaneous, liu2023translo, liu2024difflow3d, liu2025dvlo}, and virtual reality~\cite{azuma1997survey, liu2025mamba4d}. 
\cyan{The correspondence-based PCR is widely used in this field because it does not rely on initial transformation guesses ~\cite{qin2022geometric, Poiesi2021}.}
It typically consists of two main steps: (1) feature matching to establish putative 3D keypoint correspondences~\cite{choy2019fully, rusu2009fast, qin2022geometric}, and (2) \cyan{robust transformation estimation through inlier identification~\cite{zhang20233d, bai2021pointdsc, yang2020teaser}.
The high outlier ratio in real-world correspondence data makes this estimation particularly challenging.}

\begin{figure}[t!]
	\centering
	\includegraphics[width=1\linewidth, 
	trim=2mm 0mm 0mm 0mm, clip
	]{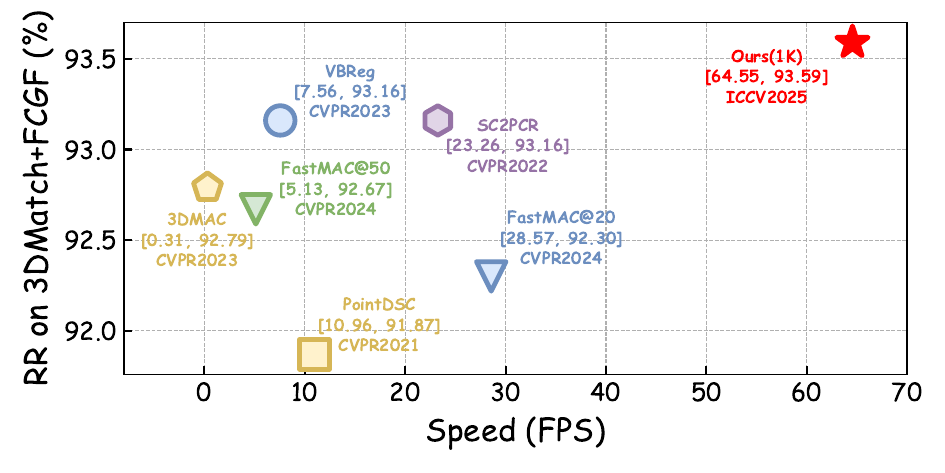}
	\vspace{-2em}
	\caption{\textbf{Registration Recall and Speed Comparison on the 3DMatch+FCGF Dataset.}  
Our method ($\textcolor{red}{\bigstar}$) achieves the highest recall and significantly outperforms competing methods in speed. 
}
	\label{fig: fig1}
	\vspace{-1em}
\end{figure}

The RANSAC family is widely used for robust PCR~\cite{fischler1981random}. These methods operate through two stages: (1) hypothesis generation and (2) model evaluation. During the hypothesis generation process, numerous putative inlier subsets are sampled from correspondences, and a rigid transformation is calculated for each subset.
In the model evaluation step, each transformation model is evaluated by metrics like inlier count, and the best transformation is finally output.
However, these methods suffer from slow convergence rates due to inefficient sampling strategies, particularly under high outlier ratios~\cite{li2023qgore}. Deep learning approaches have attempted to improve convergence efficiency through learned sampling \red{probability}~\cite{bai2021pointdsc, jiang2023robust}, but remain limited by weak generalization capabilities and substantial training requirements.

Recently, Graph-based PCR (GPCR) methods \red{gain} significant attention \red{due to their demonstrated improvements in registration robustness and accuracy.}
These methods first construct a compatibility graph based on the spatial consistency of match pairs. Then, the problem of registration becomes searching for the maximum cliques \cite{yang2020teaser, lusk2021clipper}.
SC$^2$-PCR further proposes the second-order compatibility graphs for inlier identification~\cite{Chen_2022_CVPR}. While improving registration accuracy, these methods still struggle with low inlier ratio scenarios.
\cyan{
3DMAC~\cite{zhang20233d} represents a breakthrough for low-inlier registration through maximal clique enumeration (MCE).
However, MCE has introduced two critical challenges for 3DMAC:
\redll{\textbf{(1) Inefficiency and Sensitivity:} MCE exhibits exponential time complexity with respect to correspondence \redx{number}, resulting in impractical runtime and excessive memory consumption \cite{huang2024scalable, huang2024efficient}. Although correspondence downsampling \cite{zhang2024fastmac} alleviates this inefficiency, it compromises registration accuracy. Furthermore, MCE's runtime is highly sensitive to graph density \cite{eblen2012maximum}, where sparse graphs reduce this sensitivity but simultaneously impair registration performance.}
\textbf{(2) Parallelization Limitations:} \redll{MCE algorithms face difficulties in parallel implementation due to uneven branching distributions, such as variable-sized cliques and their potentially unbounded growth \cite{almasri2023parallelizing, denacc23}.  }
These technical limitations ultimately cap how well registration systems can perform and scale.}

\redx{To address these limitations, we present TurboReg, a robust and efficient estimator for PCR.}
\redll{As illustrated in \cref{fig: fig1}, our approach achieves \redll{highest} registration recall and inference speed compared to recent robust estimators.}
The core innovation of TurboReg lies in the use of a lightweight clique to enable efficient search, combined with a highly parallelizable TurboClique \red{identification} strategy.
\redll{Specifically, we first introduce TurboClique, which is defined as a 3-clique  within a highly-constrained compatibility graph.}
The fixed-size and lightweight nature of the 3-clique provides inherent advantages in computational parallelism, while the rigorous compatibility constraints ensure that the 3-clique can estimate stable transformations by enhancing the geometry consistency between matches.
To \redx{efficiently identify TurboCliques}, 
we propose the Pivot-Guided Search (PGS) algorithm. PGS utilizes matching pairs with high $\text{SC}^2$ scores to steer the search process, thereby ensuring elevated inlier ratios for TurboCliques.
Moreover, compared to the exponential time complexity of the maximal clique search algorithm, PGS achieves linear time complexity, offering significantly higher efficiency and benefiting real-time registration.

In summary, this paper has the following contributions:

\begin{itemize}[leftmargin=2em, itemsep=0.8em, ]
    \item A novel clique structure, TurboClique, is tailored for transformation estimation, which combines a lightweight design for parallel processing with improved stability in transformation estimation.
    \item An efficient search algorithm, PGS, that identifies TurboCliques with high inlier ratios. In contrast to the exponential time complexity of MCE, PGS achieves a significantly reduced linear time complexity.
    \item The TurboReg framework, built on TurboClique and PGS, delivers state-of-the-art performance across multiple real-world datasets with significantly improved speed. For instance, on the 3DMatch+FCGF dataset, TurboReg (1K) runs $208.22\times$ faster than 3DMAC while also achieving higher recall.
\end{itemize}

\section{Related Work}
\label{sec:related}

\PAR{3D Keypoint Matching.}  Traditional 3D keypoint matching methods~\cite{aiger20084, rusu2009fast, guo2013rops, frome2004recognizing, tombari2010unique, guo2015novel, chen20073d, rusu2008aligning, zaharescu2009surface} detect reliable keypoints with descriptors and establish correspondences based on these keypoints.  
Recent advancements in this area primarily focus on learning-based feature descriptors.  
3DMatch~\cite{zeng20173dmatch}, a pioneering work, employs a Siamese network for \redll{patch-based} descriptor learning. Subsequent studies enhance performance through rotation-invariant networks~\cite{deng2018ppfnet, ao2021spinnet, wang2022you} and semantic-enhancing techniques~\cite{yan2025ml, liu2022sarnet, yin2023segregator}.  
However, those descriptors built upon sparse keypoints are at risk of losing correspondences between frames, limiting the robustness of keypoint-based methods.  
Consequently, recent approaches adopt dense matching~\cite{qin2022geometric, huang2021predator, yuan2024inlier, yu2021cofinet, yu2023rotation} to explore all potential matches in point clouds.  
Despite advancements, existing methods \redll{still face challenges due to} mismatches under extremely low inlier ratios.

\PAR{Graph-based Robust Estimators.}  
Traditional robust estimators, such as RANSAC~\cite{fischler1981random}, treat correspondences as unordered sets and rely on random subset sampling. These methods frequently exhibit slow convergence and instability under high outlier ratios. In contrast, graph-based robust estimators exploit geometric consistency to construct compatibility graphs. Approaches \redll{such as} those in~\cite{yang2023mutual, yin2023segregator, sun2022trivoc, Chen_2022_CVPR, li2023qgore} employ vote-based scoring to rank and sample correspondences. Although effective for inlier identification, these techniques lack robustness in noisy scenarios where voting scores degrade. Consequently, alternative methods adopt direct search maximum clique to maximize consensus~\cite{yang2020teaser, yin2023segregator, lusk2021clipper}. Recently, some methods relax the maximum clique to maximal clique \cite{zhang20233d, zhang2024fastmac}. \red{Although those methods achieve strong performance under low inlier ratios, they still suffer from exponential time complexity.}

\PAR{Learning-based Robust Estimators.} 
Learning-based methods aim to generate heuristic guidance for efficient \redll{matches subset sampling}, facilitating robust correspondence selection. Most existing frameworks focus on 2D image matching~\cite{yi2018learning, sun2020acne, zhang2019learning, zhao2019nm, wei2023generalized}, where the objective is to learn matching confidence scores to distinguish inliers from outliers. 
Recent 3D extensions, such as 3DRegNet~\cite{pais20203dregnet} and DGR~\cite{choy2020deep}, adopt similar pipelines: 3DRegNet uses a deep classifier, while DGR proposes a 6D convolutional U-Net for correspondence probability prediction. However, these approaches often overlook the rigid constraints inherent in 3D geometry. To address this, PointDSC~\cite{bai2021pointdsc} introduces a non-local network with an attention mechanism to enforce spatial consistency among matches, improving outlier rejection. VBReg~\cite{jiang2023robust} further refines this by integrating variational Bayesian inference to model feature uncertainty, enhancing robustness to ambiguous matches.
\redll{However, these methods rely on supervised learning, which is time-consuming to train, and struggle with generalization.}

\section{Method}
\label{sec:method}

\begin{figure}[tp]
    \centering
	\includegraphics[width=1.0\linewidth,
	]{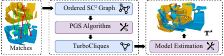}
    \caption{\textbf{Pipeline of TurboReg.} TurboReg takes correspondences as input. First, these matches are used to construct an Ordered SC$^{2}$ Graph (O2Graph, defined in \cref{def:o2graph}). Next, the PGS algorithm (\cref{sec:pgs}) is applied to the O2Graph, producing \red{TurboCliques} (defined in \cref{def: TurboClique}). Finally, during the Model Estimation step (\cref{sec:me}), a transformation is estimated for each TurboClique, and the highest-scoring transformation $\mathbf{T}^\star$ is selected to align the source and target point clouds.
}
    \label{fig:pipeline}
    \vspace{-1.em}
\end{figure}

We first outline the preliminaries of Graph-based PCR in~\cref{sec:gpcr}. \redll{Next, in~\cref{sec:TurboClique}, we explore the properties of maximal cliques, which inspire us to introduce a novel clique type called TurboClique.
We then propose an efficient TurboClique search strategy, Pivot-Guided Search (PGS) algorithm detailed in \cref{sec:pgs}.
Implementation details of PGS are discussed in \cref{sec: imp}.} Finally, we outline the model estimation process in \cref{sec:me}. The overall structure of the proposed framework is illustrated in \cref{fig:pipeline}.

\subsection{Preliminary on Graph-based PCR (GPCR) }
\label{sec:gpcr}

Given a source point cloud $\mathcal{X} \subseteq \mathbb{R}^3$ and a target point cloud $\mathcal{Y} \subseteq \mathbb{R}^3$, the goal of PCR is to compute a rigid transformation \red{$\mathbf{T} \in \text{SE}(3)$} to align them.
Matches $\mathcal{M} = \{ \boldsymbol{m}_i \}_{i=1}^N$ are first established using feature matching techniques \cite{Poiesi2021, rusu2009fast, choy2019fully}, where each correspondence $\boldsymbol{m}_i = (\boldsymbol{x}_i, \boldsymbol{y}_i)$ contains keypoints $\boldsymbol{x}_i \in \mathcal{X}$ and $\boldsymbol{y}_i \in \mathcal{Y}$. Then, a robust estimator is used to identify inliers and compute the optimal transformation~\cite{zhang20233d, Chen_2022_CVPR, yang2020teaser}. 
\redll{Our approach builds upon the recent state-of-the-art GPCR pipeline, and we outline its framework below.}

\red{The first step is to construct an undirected compatibility graph $\mathbf{G} \in \mathbb{R}^{N \times N}$ that represents the spatial compatibility of match pairs. Specifically, the $i$-th node of $\mathbf{G}$ corresponds to $\boldsymbol{m}_{i}$, and $\mathbf{G}_{ij}$ indicates whether $\boldsymbol{m}_{i}$ and $\boldsymbol{m}_{j}$ are spatially compatible, defined as:
\begin{equation}
 \mathbf{G}_{ij} = 
\begin{cases} 
1 & \text{if } \left| \Vert \boldsymbol{x}_i - \boldsymbol{x}_j \Vert_{2} - \Vert \boldsymbol{y}_i - \boldsymbol{y}_j \Vert_{2} \right| \leq \tau, \\ 
0 & \text{otherwise}, 
\end{cases}
\label{eq:ungraph}
\end{equation}
where $\tau$ denotes the compatibility threshold.}
An alternative compatibility graph is the second-order compatibility graph (SC$^2$ graph) $\hat{\mathbf{G}}$ \cite{zhang20233d, Chen_2022_CVPR}, which assigns edge weights as SC$^2$ scores:

\begin{equation}
\hat{\mathbf{G}}_{ij} = \mathbf{G}_{ij} \sum_{k=1}^N \mathbf{G}_{ik} \cdot \mathbf{G}_{jk}.
\end{equation}
Then, the maximal cliques in the compatibility are searched, and each of them is used to compute the rigid transformation~\cite{zhang20233d, qiao2023g3reg, zhang2024fastmac, wu2024quadricsreg}.
Finally, each transformation is scored using metrics like \red{inlier number}, and the transformation with the highest score is selected as the final output.

\subsection{\redx{TurboClique: Lightweight and Stable}}
\label{sec:TurboClique}

Inspired by maximal cliques, we first analyze their properties before proposing TurboClique.
Within the GPCR framework, all matches in a maximal clique estimate a rigid transformation via the Kabsch transformation solver~\cite{kabsch1976solution}. 
The stability of each estimated transformation depends on the noise distribution and the number of input matches, as the Kabsch solver relies on the least squares method.
Consider simple linear regression, where the variance of the estimated parameter $\hat{\beta}$ under Gauss-Markov assumptions is given by:
\begin{equation}
 \text{Var}(\hat{\beta}|X) = \sigma^2 (X'X)^{-1}.
\label{eq: var}
\end{equation}
Here, $\sigma$ denotes the noise level of the residuals, and $X$ represents the input observations. According to \cref{eq: var}, two primary factors determine estimation stability: (1) the residual noise $\sigma$, where lower noise reduces variance and enhances stability, and (2) the number of observation points $N$, where a larger $N$ increases $X'X$, thereby reducing variance. Consequently, employing maximal cliques to estimate transformations provides stability through two mechanisms: (1) \textbf{Data Scaling Stability:} The maximality requirement incorporates a large number of input matches, increasing the denominator of the variance term and reducing $\text{Var}(\mathbf{T})$. 
(2) \textbf{Pairwise Compatibility-induced Stability:}
Due to the inherent structure of a clique, any two matches satisfy spatial compatibility constraints, which helps mitigate random noise.
Please refer to 
\ifarxiv \cref{suppsec: pcis} \else Appendix~A.1 \fi
for details.

These two types of stability contribute to stable transformation estimation. However, the data scaling stability requires maximizing clique size, which \redx{may lead to} potentially unbounded growth \redx{of clique size} ~\cite{almasri2023parallelizing, denacc23}, resulting in significant computational and memory overhead during the search process.
This motivates the adoption of a fixed-size, lightweight clique to mitigate these limitations. Specifically, we propose using 3-clique to estimate transformation, ensuring computational efficiency while meeting the minimal requirement of transformation estimation. Although relying solely on three matches reduces data scaling stability, \redx{this limitation can be} mitigated by enhancing pairwise compatibility-induced stability.
In detail, we observe that a smaller value of $\tau$ enhances pairwise compatibility-induced stability by strengthening the spatial compatibility constraint. Therefore, we select a smaller $\tau$ to improve pairwise compatibility-induced stability, thereby compensating for the reduction in data scaling stability. Please refer to \ifarxiv \cref{suppsec: wtus} \else Appendix~A \fi for more details.
Based on these observations, we \redx{formally define} TurboClique in \cref{def: TurboClique}.

\begin{definition}[TurboClique]  
A TurboClique is a 3-clique within a compatibility graph constructed using a stringent (small) compatibility threshold.  
\label{def: TurboClique}  
\end{definition}

Notably, while a smaller $\tau$ enhances the stability of rigid transformation estimation, an excessively stringent threshold may exclude compatible inliers, \redx{thus} obscuring the distinction between inliers and outliers. Experimental results indicate that setting $\tau$ to one-fourth of the point cloud resolution strikes a reasonable balance between stability and recall. A comparison of different clique types is presented in \cref{fig: graphs}.

\subsection{Pivot-Guided Search Algorithm}
\label{sec:pgs}

\begin{figure}[tp]
	\centering
	\includegraphics[width=1.0\linewidth]{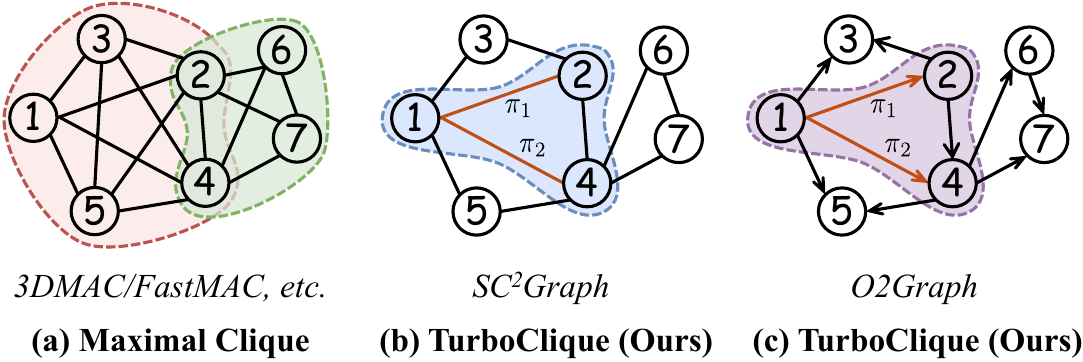}
	\caption{
\textbf{Comparison of Different Types of Cliques.} In the compatibility graph, each node represents a match, and the edges between nodes indicate spatial compatibility. Prior studies~\cite{zhang20233d, qiao2023g3reg, zhang2024fastmac, wu2024quadricsreg} primarily estimate transformation using (a) maximal clique—a subset that maximizes mutually compatible matches to ensure transformation stability. However, the computational complexity of maximal clique search grows exponentially with the size of the graph. In contrast, our proposed (b) TurboClique is lightweight while maintaining transformation stability through strict spatial compatibility constraints. 
The TurboClique on the naive SC$^2$ Graph causes redundant TurboClique detection, resulting in redundant computation. We propose a variant of the SC$^2$ Graph, (c) O2Graph, to address this issue.
}
\vspace{-0.5em}
	\label{fig: graphs}
\end{figure}

In contrast to the maximal clique, which seeks to maximize the size of the consensus subset of matches and naturally promotes a high inlier ratio, TurboClique adopts a fixed small size (only three matches). This design choice potentially limits its effectiveness in directly identifying inliers.

To address this issue, our proposed PGS algorithm leverages high-quality match pairs to guide the search process for TurboCliques. Specifically, we utilize high-scoring SC$^2$ edges, referred to as pivots, to direct the search toward TurboCliques with a high inlier ratio. Employing SC$^2$ scores as guidance is advantageous for two key reasons: (1) Higher SC$^2$ scores exhibit a strong correlation with a greater likelihood of match pairs being inliers~\cite{Chen_2022_CVPR}; (2) SC$^2$ scores reflect the density of surrounding TurboCliques—higher scores indicate a greater concentration of TurboCliques near the pivot, thus simplifying the identification of additional TurboCliques. For further details, refer to \ifarxiv \cref{suppsec: niscs} \else Appendix~B. \fi

We now formalize the PGS workflow using index notation: let $i, j, z \in \{1, \dots, N\}$ denote either graph nodes or correspondence indices. Given an SC$^2$ Graph $\hat{\mathbf{G}}$, PGS 
consists of three stages:
(\romannumeral1): Pivot Selection.  Define pivot set $\mathcal{P}$ containing the $K_{1}$ highest-weighted edges:
\begin{equation}
 \mathcal{P} = \left\{ \pi=(i, j) \, \big|\, \hat{\mathbf{G}}_{ij} \geq \alpha_{K_{1}} \right\},
\label{eq: pivot-selection}
\end{equation}
where $\alpha_{K_{1}}$ denotes the $K_{1}$-th largest edge weight.
When multiple edges share the threshold weight $\alpha_{K_{1}}$, 
\redx{all such edges are included} until $|\mathcal{P}| = K_1$.
(\romannumeral2): TurboClique Search. For each \red{pivot $\pi \in \mathcal{P}$, we find its compatible neighbors}:
\begin{equation}
 \mathcal{N}(i, j)= \left\{ z \, |\, (\hat{\mathbf{G}}_{iz} > 0) \land  ( \hat{\mathbf{G}}_{jz} > 0 )\right\}.
\label{eq: nij}
\end{equation}
Then, a TurboClique $(i,j,z)$ can be formed by combining any $z \in \mathcal{N}(i,j)$ with the pivot $(i,j)$.
However, since pivots are selected based on the SC$^2$ scores where higher scores lead to more TurboCliques, this results in uneven distribution of TurboCliques among pivots.
\red{(\romannumeral3): TurboClique Selection.
To address this imbalance, we define the aggregated weight for each TurboClique as follows:  \redx{for} all $(i, j) \in \mathcal{P}$ and $z \in \mathcal{N}(i, j)$, the aggregated weight $\mathbf{S}^{(ij)}(z)$ is given by:
 \begin{equation}
 \mathbf{S}^{(ij)}(z) = \hat{\mathbf{G}}_{ij} + \hat{\mathbf{G}}_{iz} + \hat{\mathbf{G}}_{jz}.
\label{eq: sijz}
\end{equation}
Using these aggregated weights, we retain the top-$K_2$ TurboCliques per pivot.}
Finally, we get $K_1\cdot K_2$ TurboCliques, which can be represented as:
\begin{equation}
 \mathcal{C} = \bigcup_{(i,j)\in \mathcal{P} } \left\{ (i,j,z) \,|\, z \in \underset{z\in\mathcal{N}(i,j)}{\mathrm{top}K_2}\,  \mathbf{S}^{(ij)} (z) \right\}.
\label{eq: cijz}
\end{equation}

A remaining issue is 
redundant TurboClique detection
across multiple pivots (\red{e.g., TurboClique $(1, 2,4)$ in \cref{fig: graphs}-(b) is associated with both pivots $\pi_1=(1,2)$ and $\pi_2=(1,4)$}).
To resolve this, motivated by~\cite{wang2024efficient}, we introduce a strategy of ordering the SC$^2$ graph, 
\redx{referred to as} O2Graph,  \redx{which is defined} in \cref{def:o2graph}.

\begin{definition}[Ordered SC$^2$ Graph]
	The O2Graph is a directed variant of the SC$^2$ graph $\hat{\mathbf{G}}$, denoted as $\tilde{\mathbf{G}} \in \mathbb{R}^{N \times N}$, with a key modification: edges are strictly oriented from lower-indexed to higher-indexed nodes (e.g., node $1 \to$ node $4$ in \cref{fig: graphs}-(c)), resulting in an upper triangular matrix where $\tilde{\mathbf{G}}_{ij} = 0$ for $i \geq j$. \redll{This implies that the neighbors of each node $i$ are its out-neighbors. For example, in \cref{fig: graphs}-(c), node $2$ has only nodes $4$ and $3$ as neighbors, excluding lower-indexed node $1$.}
	\label{def:o2graph}		
\end{definition}

\redx{By imposing directionality on the $\hat{\mathbf{G}}$, redundant detection of TurboCliques is avoided. Revisiting the example of TurboClique $(1, 2, 4)$ in the O2Graph (\cref{fig: graphs}-(c)), it is now exclusively associated with $\pi_2$, as node $2$ is not a neighbor of node $4$. Indeed, the O2Graph rigorously ensures that each TurboClique is detected by at most one pivot, a property we term the \textit{Unique Assignment Property of TurboClique}.} Please refer to \ifarxiv \cref{supsec: unique} \else Appendix~C \fi for more details.

\subsection{Implemetation Details of PGS}
\label{sec: imp}

This section outlines the efficient implementation of the PGS algorithm. The pseudo-code for PGS, presented in Algorithm~\ref{algo:pgs-general}, accepts either an SC$^2$ Graph or an O2Graph as input, denoted as $\bar{\mathbf{G}} \in \{ \hat{\mathbf{G}}, \tilde{\mathbf{G}} \}$.
The computational complexity primarily stems from $K_{1}$ iterations (Line~5) , \redx{each involving up} to $N-2$ neighbor identifications (Line~8), resulting in a total computational workload of $\mathcal{O}(K_{1} N)$. 
Note that $K_{1}$ is a user-defined hyperparameter independent of $N$. Therefore, the overall time complexity remains linear, i.e., $\mathcal{O}(N)$.
Practically, all $K_{1}(N-2)$ TurboClique searches exhibit two levels of parallelism: 
(1) \textbf{Pivot-level Parallelism}: Independent processing of each pivot enables parallel of the main loop (Line~5); (2) \textbf{Search-level Parallelism}: Concurrent execution of TurboClique searches within each pivot iteration (Line~8).
Furthermore, for any $(i, j) \in \mathcal{P}$ and $z \in \{1, \dots, N\} \setminus \{i, j\}$, the verification condition \redx{determining} whether $(i,j,z)$ is a TurboClique can be simply represented as:
\begin{equation}
\bar{\mathbf{G}}_{ij} \cdot \bar{\mathbf{G}}_{iz} \cdot \bar{\mathbf{G}}_{jz} > 0.
\end{equation}
This implies that the TurboClique search can be formulated as a dense matrix element-wise multiplication problem, which is inherently suitable for efficient GPU implementation. Leveraging the aforementioned two levels of parallelism, where all $K_{1}(N-2)$ TurboClique computations can be executed in parallel, the time complexity on a GPU reduces to $\mathcal{O}\left(\frac{K_1 N}{R}\right)$, with $R$ being the number of parallel processing units. In scenarios where $R \gg K_1 N$, the GPU implementation effectively achieves near-constant time complexity, approximating $\mathcal{O}(1)$.  
Please refer to \ifarxiv \cref{appsec:code} \else Appendix~D \fi
for Tensor-style pseudo-code.

Notably, PGS significantly outperforms the MCE algorithm used in 3DMAC~\cite{zhang20233d}, which has a complexity of $\mathcal{O}\left(d(N-d)3^{d/3}\right)$, where $d$ is the graph's degeneracy.

\begin{algorithm}[t!]
	\SetAlgoLined	
	\textbf{Input: } Weighted graph: $\bar{\mathbf{G}} \in \mathbb{R}^{N \times N}$; number of pivots $K_1 \in \mathbb{N}^+$; number of TurboCliques for each pivot $K_2 \in \mathbb{N}^+$  \\
	\textbf{Output: } TurboClique set $\mathcal{C}$  \\
	$\mathcal{C} \gets \emptyset$ \\
	$\mathcal{P} \gets \text{TopKEdges}(\bar{\mathbf{G}}, K_1) \quad$ \annot{\cref{eq: pivot-selection}}  \\
	\For{$(i, j) \in \mathcal{P}$} {
		 $\mathcal{N}(i, j) \gets \emptyset$ \\
		 $\mathbf{S}^{(ij)} \gets \{ 0 \}^{N}$ \\
		\For{ $z \in \{ 1, \dots, N \} \setminus \{ i, j \}$  } {
			\If {$(\bar{\mathbf{G}}_{iz} > 0) \land  ( \bar{\mathbf{G}}_{jz} > 0 )$}    {
				$\mathbf{S}^{(ij)}(z)=\bar{\mathbf{G}}_{ij}+\bar{\mathbf{G}}_{iz}+\bar{\mathbf{G}}_{jz} \quad$   \annot{\cref{eq: sijz}} \\
			}
		} 
		$\mathcal{Z}^{\text{top}} \gets \text{SelectTopK}(\mathbf{S}^{ij}, K_{2})\quad$ \annot{\cref{eq: cijz}} \\
		\For{$z \in \mathcal{Z}^{\text{top}}$} {
			$\mathcal{C} \gets \mathcal{C} \cup \{(i, j, z)\} \quad$  \annot{\cref{eq: cijz}} \\
		}
	}
	\Return $\mathcal{C}$
	\caption{Pivot-Guided Search Algorithm}
    \label{algo:pgs-general}
\end{algorithm}

\subsection{Model Estimation}
\label{sec:me}

Finally, the Kabsch pose solver~\cite{kabsch1976solution} is used to estimate transformation for each TurboClique,  denoted as $\mathcal{T} = \{ \mathbf{T}_{z} \}_{z=1}^{K_{1} \cdot  K_{2}}$. Then, the optimal rigid transformation is selected by:
\begin{equation}
\mathbf{T}^{\star} = \underset{\mathbf{T}_{z} \in \mathcal{T} }{\text{argmax}} \ g(\mathbf{T}_{z}),
\end{equation}
where $g(\cdot)$ quantifies the inlier number (IN) of the rigid transformation $\mathbf{T}_{z}$.

\section{Experiments}
\label{sec:exp}

\redx{In this section, we first validate the performance of TurboReg on both indoor and outdoor datasets in \cref{sec:reg_exp}, demonstrating its robustness and effectiveness across diverse environments. 
We then conduct runtime distribution analysis experiments to assess the efficiency and low temporal variability of our algorithm in \cref{sec: time}.
Finally, in \cref{exp:analysis}, we perform a detailed parameter ablation study to offer deeper insights into the key factors that influence performance.}

\subsection{Registration Experiments}
\label{sec:reg_exp}

\subsubsection{Experimental Setup}

\newcommand{\ourszf}{Ours (0.5K) }
\newcommand{\oursztf}{Ours (0.25K) }
\newcommand{\ourso}{Ours (1K) }
\newcommand{\oursof}{Ours (1.5K) }
\newcommand{\ourst}{Ours (2K) }
\newcommand{\ourstf}{Ours (2.5K) }

\PAR{Datasets.} 
We follow dataset settings of \cite{zhang20233d, Chen_2022_CVPR, jiang2023robust} to evaluate our method on both indoor and outdoor datasets. Specifically, for indoor scene registration, we use the 3DMatch~\cite{zeng20173dmatch} and 3DLoMatch~\cite{huang2021predator} datasets. 
The 3DMatch dataset consists of 1623 point cloud pairs, while 3DLoMatch contains 1781 pairs with a more challenging overlap rate between 10\% and 30\%. 
We employ traditional descriptors such as FPFH~\cite{rusu2009fast}, as well as deep learning-based descriptors like FCGF~\cite{choy2019fully} and Predator~\cite{huang2021predator} for feature matching. For outdoor scene registration, we use the KITTI~\cite{Geiger2012CVPR} dataset, which includes 555 point cloud pairs. Similar to the indoor datasets, we utilize FPFH~\cite{rusu2009fast} and FCGF~\cite{choy2019fully} descriptors for feature matching, in line with previous studies~\cite{zhang20233d, Chen_2022_CVPR}. 


\PAR{Implementation Details.} Our experimental setup employs an Intel i7-13700KF CPU and an NVIDIA RTX 4090 GPU. 
TurboReg is implemented in C++ using the LibTorch library \cite{paszke2019pytorch}. 
Additionally, we provide a CPU implementation version of TurboReg, which utilizes the Eigen library. 
The compatibility threshold is initially set to $0.25 \times \text{pr}$ (point cloud resolution \cite{zhang20233d}) and adjusted based on empirical evaluation.
%
We set the number of pivots $K_1$ to 1K and 2K for indoor datasets, and to 0.25K and 0.5K for the outdoor dataset.  
The number of TurboCliques per pivot is fixed at $K_2 = 2$.

\PAR{Baselines.} We evaluate our method against a diverse set of baselines, encompassing both deep learning-based and traditional approaches. The deep learning baselines include DGR~\cite{choy2020deep}, VBReg~\cite{jiang2023robust}, and PointDSC~\cite{bai2021pointdsc}.
For traditional methods, we consider RANSAC~\cite{fischler1981random}, GC-RANSAC~\cite{barath2018graph}, TEASER++~\cite{yang2020teaser}, CG-SAC~\cite{quan2020compatibility}, SC$^2$-PCR~\cite{Chen_2022_CVPR}, 3DMAC~\cite{zhang20233d}, and FastMAC~\cite{zhang2024fastmac} with varying sample ratios (e.g., FastMAC@20 and FastMAC@50 correspond to sample ratios of 20\% and 50\%, respectively).

\PAR{Metrics.} We report the rotation error (RE) and translation error (TE) to evaluate registration accuracy~\cite{choy2019fully, huang2021predator, liu2023regformer}. Following prior works~\cite{zhang20233d, Chen_2022_CVPR}, a registration is considered successful if RE $\leq 15^\circ$ and TE $\leq 30$ cm for the 3DMatch and 3DLoMatch datasets, and if RE $\leq 5^\circ$ and TE $\leq 60$ cm for the KITTI dataset.  
We also compute the registration recall (RR)~\cite{huang2021predator}, defined as the ratio of successful registrations (RE $\le 15^\circ$, TE $\le 30$~cm) to the total number of point cloud pairs. Additionally, we report
the speed of methods in terms of frames per second (FPS).
\red{For datasets using both FPFH and FCGF descriptors, we follow ~\cite{zhang2024fastmac} and report the average computation time of the two descriptor types.}

\subsubsection{Indoor Registration}
\label{sec:pcr_indoor}

\begin{table}[t]
	\centering
	\renewcommand{\arraystretch}{1}
	\renewcommand\tabcolsep{1.5pt}
	\resizebox{1.0\columnwidth}{!}{
		\begin{tabular}{l|ccc|ccc|cc}
			\toprule
			\midrule
			\multirow{2}[2]{*}{Methods}          & \multicolumn{3}{c|}{FPFH}  & \multicolumn{3}{c|}{FCGF}           & \multicolumn{2}{c}{FPS}                                                                                                                                             \\
			                                     & \multicolumn{1}{l}{RR(\%)} & \multicolumn{1}{l}{RE(\textdegree)} & \multicolumn{1}{l|}{TE(cm)} & \multicolumn{1}{l}{RR(\%)} & \multicolumn{1}{l}{RE(\textdegree)} & \multicolumn{1}{l|}{TE(cm)} & CPU              & GPU               \\
			\midrule
			\emph{\romannumeral1)~Deep Learned}  &                            &                                     &                             &                            &                                     &                             &                  &                   \\
			DGR~\cite{choy2020deep}              & 32.84                      & 2.45                                & 7.53                        & 88.85                      & 2.28                                & 7.02                        & 0.43             & 0.91              \\
			PointDSC~\cite{bai2021pointdsc}      & 72.95                      & {2.18}                              & \textbf{6.45}               & 91.87                      & {2.10}                              & {6.54}                      & 0.20             & 10.74             \\
			VBReg~\cite{jiang2023robust}         & 82.75                      & \underline{2.14}                    & 6.77                        & \textbf{93.16}             & 2.33                                & 6.68                        & 0.06             & 7.62              \\
			\midrule
			\emph{\romannumeral2)~Learning Free} &                            &                                     &                             &                            &                                     &                             &                  &                   \\
			RANSAC-1M~\cite{fischler1981random}  & 64.20                      & 4.05                                & 11.35                       & 88.42                      & 3.05                                & 9.42                        & 0.05             & -                 \\
			RANSAC-4M~\cite{fischler1981random}  & 66.10                      & 3.95                                & 11.03                       & 91.44                      & 2.69                                & 8.38                        & 0.01             & -                 \\
			GC-RANSAC~\cite{barath2018graph}     & 67.65                      & 2.33                                & 6.87                        & 92.05                      & 2.33                                & 7.11                        & 1.01             & -                 \\
			TEASER++~\cite{yang2020teaser}       & 75.48                      & 2.48                                & 7.31                        & 85.77                      & 2.73                                & 8.66                        & 1.12             & -                 \\
			CG-SAC~\cite{quan2020compatibility}  & 78.00                      & 2.40                                & 6.89                        & 87.52                      & 2.42                                & 7.66                        & -                & -                 \\
			$\rm{SC}^2$-PCR~\cite{chen2022sc2}   & 83.73                      & {2.18}                              & \underline{6.70}            & \textbf{93.16}             & 2.09                                & {6.51}                      & 0.27             & 15.84             \\
			3DMAC~\cite{zhang20233d}             & \underline{83.92}          & \textbf{2.11}                       & 6.80                        & 92.79                      & 2.18                                & 6.89                        & 0.31             & -                 \\
			FastMAC@50~\cite{zhang2024fastmac}   & 82.87                      & 2.15                                & {6.73}                      & 92.67                      & \textbf{2.00}                       & 6.47                        & 1.35             & 4.33              \\
			FastMAC@20~\cite{zhang2024fastmac}   & 80.71                      & 2.17                                & 6.81                        & 92.30                      & \underline{2.02}                    & 6.52                        & 1.56             & 26.32             \\
			\midrule
			\ourso                               & \underline{83.92}          & 2.17                                & 6.79                        & \textbf{93.59}             & 2.03                                & \textbf{6.40}               & \textbf{2.73}    & \textbf{61.25}    \\
			\ourst                               & \textbf{84.10}             & 2.19                                & 6.81                        & \textbf{93.59}             & 2.04                                & \underline{6.42}            & \underline{2.46} & \underline{54.04} \\
			\midrule
			\bottomrule
		\end{tabular}}
	\vspace{-0.3em}
	\caption{Registration results on 3DMatch dataset.}
	\label{tab: exp_3dmatch_fpfh_fcgf}
	\vspace{-0.5em}
\end{table}

\begin{table}[t]
	\centering
	\renewcommand{\arraystretch}{1}
	\renewcommand\tabcolsep{1.5pt}
	\resizebox{1.0\columnwidth}{!}{
		\begin{tabular}{l|ccc|ccc|cc}
			\toprule
			\midrule
			\multirow{2}[2]{*}{Methods}          & \multicolumn{3}{c|}{FPFH}  & \multicolumn{3}{c|}{FCGF}           & \multicolumn{2}{c}{FPS}                                                                                                                                           \\
			                                     & \multicolumn{1}{l}{RR(\%)} & \multicolumn{1}{l}{RE(\textdegree)} & \multicolumn{1}{l|}{TE(cm)} & \multicolumn{1}{l}{RR(\%)} & \multicolumn{1}{l}{RE(\textdegree)} & \multicolumn{1}{l|}{TE(cm)} & CPU             & GPU              \\
			\midrule
			\emph{\romannumeral1)~Deep Learned}  &                            &                                     &                             &                            &                                     &                             &                 &                  \\
			DGR~\cite{choy2020deep}              & 19.88                      & \textbf{5.07}                       & 13.53                       & 43.80                      & 4.17                                & 10.82                       & 0.48            & 1.01             \\
			PointDSC~\cite{bai2021pointdsc}      & 20.38                      & 4.04                                & 10.25                       & 56.20                      & 3.87                                & 10.48                       & 0.20            & 10.20            \\
			\midrule
			\emph{\romannumeral2)~Learning Free} &                            &                                     &                             &                            &                                     &                             &                 &                  \\
			RANSAC-1M~\cite{fischler1981random}  & 0.67                       & 10.27                               & 15.06                       & 9.77                       & 7.01                                & 14.87                       & 0.05            & -                \\
			RANSAC-4M~\cite{fischler1981random}  & 0.45                       & 10.39                               & 20.03                       & 10.44                      & 6.91                                & 15.14                       & 0.01            & -                \\
			TEASER++~\cite{yang2020teaser}       & 35.15                      & 4.38                                & 10.96                       & 46.76                      & 4.12                                & 12.89                       & 1.26            & -                \\
			$\rm{SC}^2$-PCR~\cite{chen2022sc2}   & {38.57}                    & 4.03                                & 10.31                       & 58.73                      & 3.80                                & 10.44                       & 0.28            & 12.82            \\
			3DMAC~\cite{zhang20233d}             & \underline{40.88}          & \textbf{3.66}                       & \textbf{9.45}               & \textbf{59.85}             & \textbf{3.50}                       & \textbf{9.75}               & 0.32            & -                \\
			FastMAC@50~\cite{zhang2024fastmac}   & 38.46                      & 4.04                                & 10.47                       & 58.23                      & 3.80                                & 10.81                       & 0.27            & 5.05             \\
			FastMAC@20~\cite{zhang2024fastmac}   & 34.31                      & 4.12                                & 10.82                       & 55.25                      & 3.84                                & 10.71                       & 1.28            & 20.41            \\
			\midrule
			\ourso                               & 40.76                      & 3.91                                & 10.23                       & 59.40                      & \underbar{3.72}                     & \underbar{10.31}            & \textbf{2.88}   & \textbf{61.87}   \\
			\ourst                               & \textbf{40.99}             & \underbar{3.85}                     & \underline{10.16}           & \underbar{59.74}           & 3.76                                & 10.41                       & \underbar{2.84} & \underbar{51.59} \\
			\midrule
			\bottomrule
		\end{tabular}}
	\vspace{-0.3em}
	\caption{Registration results on 3DLoMatch dataset.}
	\label{tab: 3dlomatch_fpfh_fcgf}
	\vspace{-0.5em}
\end{table}%

\begin{table}[t]
	\centering
	\renewcommand{\arraystretch}{1}
	\renewcommand\tabcolsep{7pt}
	\resizebox{1.0\columnwidth}{!}{
		\begin{tabular}{l|c|cc|c|cc}
			\toprule
			\midrule
			\multirow{3}[3]{*}{Estimator}       & \multicolumn{3}{c|}{3DMatch} & \multicolumn{3}{c}{3DLoMatch}                                                                                            \\ \cline{2-7}
			                                    & \multirow{2}{*}{RR(\%)}      & \multicolumn{2}{c|}{FPS}      & \multirow{2}{*}{RR(\%)} & \multicolumn{2}{c}{FPS}                                        \\
			                                    &                              & CPU                           & GPU                     &                         & CPU              & GPU               \\

			\midrule
			RANSAC-1M~\cite{fischler1981random} & 89.23                        & 0.05                          & -                       & 54.97                   & 0.05             & -                 \\
			RANSAC-4M~\cite{fischler1981random} & 91.72                        & 0.01                          & -                       & 62.88                   & 0.01             & -                 \\
			TEASER++~\cite{yang2020teaser}      & 93.16                        & 1.14                          & -                       & 64.07                   & 1.15             & -                 \\
			SC$^2$-PCR~\cite{chen2022sc2}       & 93.59                        & 0.31                          & 13.99                   & 69.57                   & 0.30             & 14.72             \\
			3DMAC~\cite{zhang20233d}            & \underline{94.60}            & 0.41                          & -                       & 70.90                   & 0.59             & -                 \\
			FastMAC@50~\cite{zhang2024fastmac}  & 93.72                        & 1.33                          & 8.84                    & 69.12                   & 2.12             & 13.38             \\
			FastMAC@20~\cite{zhang2024fastmac}  & 93.10                        & 1.57                          & 25.67                   & 68.50                   & 2.34             & 29.31             \\
			\midrule
			\ourso                              & \textbf{94.89}               & \textbf{3.56}                 & \textbf{61.93}          & \textbf{73.07}          & \textbf{4.52}    & \textbf{72.73}    \\
			\ourst                              & \underline{94.60}            & \underline{2.81}              & \underline{61.34}       & \underline{72.95}       & \underline{3.73} & \underline{55.87} \\
			\midrule
			\bottomrule
		\end{tabular}}
	\caption{Registration results on the 3DMatch and 3DLoMatch datasets using the Predator descriptor.}
	\label{tab: 3dlomatch_predator}
	\vspace{-1.5em}
\end{table}

\PAR{Results on the 3DMatch.}  
The registration results on the 3DMatch dataset, presented in \cref{tab: exp_3dmatch_fpfh_fcgf}, demonstrate that our method achieves state-of-the-art RR, reaching 84.10\% with FPFH (0.18\% above 3DMAC~\cite{zhang20233d}) and 93.59\% with FCGF.
These findings confirm two key points: (1) a 3-clique suffices for effective registration, and (2) TurboReg leverages the PGS algorithm to efficiently detect cliques containing inliers.
Note that while our method achieves the highest performance on RR, its RE and TE are slightly higher. Since our method achieves higher recall by incorporating more challenging cases, this consequently increases RE and TE.

Runtime performance positions Ours (1K) and \ourst as the top two fastest methods across both CPU and GPU platforms.
On the GPU, all configurations deliver real-time performance: 
Compared to FastMAC@50, \ourst achieves a $12.48 \times$ speedup on GPU and a $1.82 \times$ speedup on CPU.
This efficiency stems from the highly parallelizable design of the PGS algorithm and its efficient implementation.

\PAR{Results on the 3DLoMatch Dataset.} The registration results for the 3DLoMatch dataset are presented in \cref{tab: 3dlomatch_fpfh_fcgf}. TurboReg achieves state-of-the-art RR while delivering substantially higher speeds. For example, when using the FPFH descriptor, TurboReg (2K) outperforms 3DMAC in RR, with speed improvements of $8.88\times$ on CPU and $161.22\times$ on GPU. 
\ourst achieves a $10\times$ speedup over FastMAC@50, while improving RR by 2.53\% (FPFH) and 1.51\% (FCGF), highlighting TurboReg's efficiency and robustness in low-overlap scenarios.

\PAR{Registration Results with Predator.} 
Experiments with the Predator descriptor on 3DMatch and 3DLoMatch datasets are reported in \cref{tab: 3dlomatch_predator}. 
Results show that TurboReg (1K) achieves the highest RR, with 94.89\% on 3DMatch (0.29\% above 3DMAC) and 73.07\% on 3DLoMatch (2.17\% above 3DMAC), while also delivering the fastest runtime.

\subsubsection{Outdoor Registration}
\label{sec:pcr_outdoor}
The registration results on the KITTI dataset are presented in \cref{tab:kitti}. 
TurboReg achieves the best performance in both registration recall and speed. 
For instance, \oursztf achieves RR values of 98.56\% and 98.38\% on the FPFH and FCGF descriptors, respectively, surpassing all baselines. 
Notably, \redll{Ours (0.25K)} generates only 500 pivots, demonstrating its effectiveness in the hypothesis generation, which is significantly lower than the number of RANSAC iterations, which require millions of hypotheses.

\begin{table}[t]
	\centering
	\renewcommand{\arraystretch}{1}
	\renewcommand\tabcolsep{1.5pt}
	\resizebox{1.0\columnwidth}{!}{
		\begin{tabular}{l|ccc|ccc|cc}
			\toprule
			\midrule
			\multirow{2}[2]{*}{Methods}          & \multicolumn{3}{c|}{FPFH}  & \multicolumn{3}{c|}{FCGF}           & \multicolumn{2}{c}{FPS}                                                                                                                                             \\
			                                     & \multicolumn{1}{l}{RR(\%)} & \multicolumn{1}{l}{RE(\textdegree)} & \multicolumn{1}{l|}{TE(cm)} & \multicolumn{1}{l}{RR(\%)} & \multicolumn{1}{l}{RE(\textdegree)} & \multicolumn{1}{l|}{TE(cm)} & CPU              & GPU               \\
			\midrule
			\emph{\romannumeral1)~Deep Learned}  &                            &                                     &                             &                            &                                     &                             &                  &                   \\
			DGR~\cite{choy2020deep}              & 77.12                      & 1.64                                & 33.10                       & 94.90                      & \textbf{0.34}                       & 21.70                       & 0.41             & 1.02              \\
			PointDSC~\cite{bai2021pointdsc}      & 96.40                      & \textbf{0.38}                       & \underline{8.35}            & 96.40                      & 0.61                                & 13.42                       & 0.19             & 8.99              \\ \hline
			\emph{\romannumeral2)~Learning Free} &                            &                                     &                             &                            &                                     &                             &                  &                   \\
			TEASER++~\cite{yang2020teaser}       & 91.17                      & 1.03                                & 17.98                       & 94.96                      & 0.38                                & 13.69                       & 1.13             & -                 \\
			RANSAC-4M~\cite{fischler1981random}  & 74.41                      & 1.55                                & 30.20                       & 80.36                      & 0.73                                & 26.79                       & 0.01             &                   \\
			CG-SAC~\cite{quan2020compatibility}  & 74.23                      & 0.73                                & 14.02                       & 83.24                      & 0.56                                & 22.96                       & -                & -                 \\
			$\rm{SC}^2$-PCR~\cite{chen2022sc2}   & 96.40                      & \underline{0.41}                    & \textbf{8.00}               & 97.12                      & 0.41                                & 9.71                        & 0.31             & 14.03             \\
			3DMAC~\cite{zhang20233d}             & 97.66                      & \underline{0.41}                    & 8.61                        & 97.25                      & \underline{0.36}                    & 8.00                        & 0.34             & -                 \\
			FastMAC@50~\cite{zhang2024fastmac}   & 97.84                      & \underline{0.41}                    & 8.61                        & 97.84                      & \underline{0.36}                    & \underline{7.98}            & 1.40             & 9.26              \\
			FastMAC@20~\cite{zhang2024fastmac}   & \underline{98.02}          & \underline{0.41}                    & 8.64                        & 97.48                      & 0.38                                & 8.20                        & 1.45             & 34.48             \\
			\midrule
			\oursztf                             & \textbf{98.56}             & 0.47                                & 8.96                        & \textbf{98.38}             & 0.40                                & 8.12                        & \textbf{2.55}    & \textbf{61.00}    \\
			\ourszf                              & 97.84                      & 0.46                                & 8.68                        & \underline{98.20}          & 0.39                                & \textbf{7.91}               & \underline{2.12} & \underline{58.92} \\
			\midrule
			\bottomrule
		\end{tabular}}
	\caption{Registration results on KITTI dataset.}
	\vspace{-1em}
	\label{tab:kitti}
\end{table}

\subsection{Runtime Distribution Analysis}
\label{sec: time}

\begin{figure}[tp]
	\centering
	\includegraphics[width=\linewidth,
	trim=2.7mm 2mm 4mm 1mm, clip 
	]{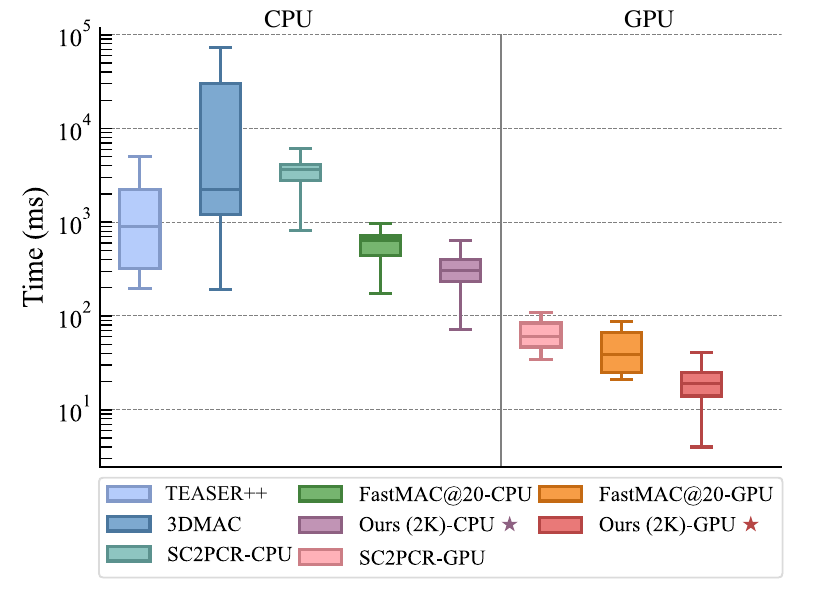}
	\vspace{-2em}
	\caption{Time comparison (ms) between various robust estimators, with implementations on both CPU and GPU. \red{Our methods (\textcolor{mygreen}{$\bigstar$} and \textcolor{myorange}{$\bigstar$}) achieve the fastest speeds and most stable runtime distribution on both CPU and GPU platforms.}
}
	\label{fig: time_copa}
	\vspace{-1em}
\end{figure}

\cref{fig: time_copa} presents a comparative analysis of computational efficiency across state-of-the-art robust estimators on the 3DMatch+FPFH benchmark, with box plots illustrating runtime distributions (in milliseconds) for both CPU and GPU implementations.  Our method demonstrates a statistically significant reduction in computational latency compared to competitors on both CPU and GPU, with a tightly clustered distribution indicating superior temporal stability.

\begin{table*}[t!]
	\centering
	\renewcommand{\arraystretch}{1}
	\renewcommand\tabcolsep{6.2pt}
	\resizebox{2.05\columnwidth}{!}{
		\begin{tabular}{l|c|l|ccccc|ccccc|ccccc}
			\toprule
			\midrule
			 &     & \multirow{3}{*}{Parameters}     & \multicolumn{5}{c|}{3DMatch+FPFH} & \multicolumn{5}{c|}{3DMatch+FCGF} & \multicolumn{5}{c}{3DLoMatch+Predator}                                                                                                                                                                                                                                                                                                                         \\ \cline{4-18}
			 &     &                                 & \multirow{2}{*}{RR(\%)}           & \multirow{2}{*}{RE(\textdegree)}  & \multirow{2}{*}{TE(cm)}                & \multicolumn{2}{c|}{FPS} & \multirow{2}{*}{RR(\%)} & \multirow{2}{*}{RE(\textdegree)} & \multirow{2}{*}{TE(cm)} & \multicolumn{2}{c|}{FPS} & \multirow{2}{*}{RR(\%)} & \multirow{2}{*}{RE(\textdegree)} & \multirow{2}{*}{TE(cm)} & \multicolumn{2}{c}{FPS}                                                           \\
			 &     &                                 &                                   &                                   &                                        & CPU                      & GPU                     &                                  &                         &                          & CPU                     & GPU                              &                         &                         &                  & CPU              & GPU               \\
			\midrule
			\multirow{10}{*}{\rotatebox{90}{O2Graph Construction}}
			 & 1)  & $\tau=0.0005$ (m)               & 66.30                             & \textbf{1.84}                     & \textbf{5.99}                          & \textbf{9.12}            & 47.25                   & 92.05                            & \textbf{1.98}           & 6.35                     & \textbf{8.12}           & 57.16                            & 62.85                   & 3.28                    & \underline{9.32} & \textbf{11.23}   & 71.32             \\
			 & 2)  & $\tau=0.001$ (m)                & 76.96                             & 2.05                              & 6.47                                   & \underline{5.48}         & 50.16                   & 93.04                            & \underline{2.01}        & 6.38                     & \underline{5.66}        & 61.28                            & 69.62                   & 3.32                    & 9.47             & \underline{8.34} & 64.90             \\
			 & 3)  & $\tau=0.010$ (m)                & 83.61                             & 2.14                              & 6.68                                   & 3.12                     & 52.59                   & 93.28                            & 2.02                    & 6.35                     & 4.01                    & 62.96                            & \textbf{73.07}          & 3.28                    & 9.53             & 4.36             & 67.68             \\
			 & 4)  & $\tau=0.012$ (m)                & \textbf{84.10}                    & 2.19                              & 6.81                                   & 2.49                     & 52.87                   & \textbf{93.59}                   & 2.03                    & 6.40                     & 2.74                    & 64.55                            & 72.83                   & \underline{3.27}        & 9.45             & 4.69             & 72.60             \\
			 & 5)  & $\tau=0.014$ (m)                & 83.73                             & 2.14                              & 6.67                                   & 2.12                     & 52.29                   & 93.28                            & 2.02                    & 6.39                     & 3.87                    & 64.50                            & 72.89                   & 3.30                    & 9.50             & 4.31             & 72.50             \\
			 & 6)  & $\tau=0.04$ (m)                 & 82.93                             & 2.12                              & 6.68                                   & 1.51                     & 50.31                   & 92.73                            & \underline{2.01}        & 6.35                     & 1.67                    & 58.93                            & 71.84                   & 3.29                    & 9.40             & 1.45             & 70.96             \\
			 & 7)  & $\tau=0.08$ (m)                 & 82.44                             & 2.09                              & 6.63                                   & 1.37                     & 49.37                   & 92.67                            & \underline{2.01}        & \underline{6.33}         & 1.45                    & 58.86                            & 70.73                   & 3.29                    & 9.35             & 1.11             & 69.47             \\
			 & 8)  & $\tau=0.2$ (m)                  & 81.33                             & 2.12                              & 6.56                                   & 0.77                     & 47.26                   & 91.68                            & 2.02                    & 6.38                     & 0.58                    & 53.01                            & 65.25                   & 3.35                    & 9.35             & 0.59             & 61.47             \\
			 & 9)  & $\tau=0.5$ (m)                  & 74.43                             & \underline{2.04}                  & \underline{6.33}                       & 0.76                     & 47.36                   & 86.69                            & \underline{2.01}        & \textbf{6.31}            & 0.43                    & 49.17                            & 53.05                   & \textbf{3.20}           & \textbf{9.06}    & 0.47             & 59.01             \\
			\cline{2-18}
			 & 10) & SC$^2$ Graph $\hat{\mathbf{G}}$ & 83.61                             & 2.16                              & 6.82                                   & 2.13                     & 49.44                   & \textbf{93.59}                   & 2.04                    & 6.42                     & 2.37                    & 59.12                            & 72.89                   & 3.35                    & 9.56             & 3.99             & 56.21             \\
			\midrule

			\multirow{10}{*}{\rotatebox{90}{PGS Algorithm}}
			 & 11) & $K_1=10$                        & 78.68                             & 2.10                              & 6.49                                   & 3.24                     & 63.21                   & 91.74                            & \underline{2.01}        & 6.38                     & 3.33                    & 62.34                            & 69.50                   & 3.35                    & 9.49             & 4.92             & \underline{73.74} \\
			 & 12) & $K_1=50$                        & 82.19                             & 2.12                              & 6.70                                   & 3.22                     & 59.23                   & 92.42                            & \underline{2.01}        & 6.36                     & 3.33                    & 62.15                            & 71.84                   & 3.32                    & 9.44             & 4.92             & 69.77             \\
			 & 13) & $K_1=250$                       & 83.12                             & 2.15                              & 6.72                                   & 3.23                     & 62.50                   & 93.28                            & 2.02                    & 6.37                     & 3.31                    & \underline{65.93}                & \underline{72.95}       & 3.29                    & 9.55             & 4.88             & 70.05             \\
			 & 14) & $K_1=500$                       & 83.36                             & 2.15                              & 6.76                                   & 3.20                     & \underline{63.12}       & 93.28                            & \underline{2.01}        & 6.37                     & 3.19                    & \textbf{66.14}                   & \textbf{73.07}          & 3.28                    & 9.53             & 4.52             & 72.73             \\
			 & 15) & $K_1=1000$                      & \underline{83.92}                 & 2.17                              & 6.79                                   & 2.71                     & 58.12                   & \textbf{93.59}                   & 2.03                    & 6.40                     & 2.74                    & 64.55                            & \textbf{73.07}          & 3.30                    & 9.52             & 4.36             & 67.68             \\
			 & 16) & $K_1=2000$                      & \textbf{84.10}                    & 2.19                              & 6.81                                   & 2.49                     & 52.87                   & 93.22                            & 2.02                    & 6.38                     & 2.42                    & 55.21                            & \underline{72.95}       & 3.29                    & 9.50             & 3.73             & 55.87             \\

			\cline{2-18}
			 & 17) & $K_2=1$                         & 83.43                             & 2.06                              & 6.58                                   & 2.84                     & \textbf{63.13}          & 93.22                            & 2.02                    & 6.35                     & 3.12                    & 65.33                            & 72.89                   & 3.41                    & {9.64}           & 4.83             & \textbf{78.36}    \\
			 & 18) & $K_2=2$                         & \textbf{84.10}                    & 2.19                              & 6.81                                   & 2.49                     & 52.87                   & \textbf{93.59}                   & 2.03                    & 6.40                     & 2.74                    & 64.55                            & \textbf{73.07}          & 3.28                    & 9.53             & 4.36             & 67.68             \\
			 & 19) & $K_2=3$                         & \textbf{84.10}                    & 2.19                              & 6.81                                   & 1.98                     & 46.30                   & \underline{93.53}                & 2.03                    & 6.45                     & 2.32                    & 61.31                            & \textbf{73.07}          & 3.30                    & 9.53             & 3.98             & 68.98             \\
			\midrule
			 & 23) & Default                         & \textbf{84.10}                    & 2.19                              & 6.81                                   & 2.49                     & 52.87                   & \textbf{93.59}                   & 2.03                    & 6.40                     & 2.74                    & 64.55                            & \textbf{73.07}          & 3.28                    & 9.53             & 4.36             & 67.68             \\
			\midrule
			\bottomrule
		\end{tabular}
	}
	\caption{Ablation study on 3DMatch and 3DLoMatch datasets.}
	\label{table: Ablation_study}
	\vspace{-1.5em}
\end{table*}

\subsection{Ablation Study}
\label{exp:ablation}

\label{exp:analysis}

We conduct ablation studies across three configurations: 3DMatch+FPFH, 3DMatch+FCGF, and 3DLoMatch+Predator. 
The results are presented in \cref{table: Ablation_study}. 
\redx{We divide the configurations into two groups based on the key steps of TurboReg:}
(1) O2Graph Construction: We evaluate compatibility thresholds $\tau$ ranging from $0.0005$~m to $0.5$~m. 
Our default compatibility graph is O2Graph $\tilde{\mathbf{G}}$, which we compare with the undirected SC$^2$ graph $\hat{\mathbf{G}}$ (Line~10).
(2) PGS Algorithm: \redx{We vary $K_1$ (pivot count) from 10 to 2000 and $K_2$ (TurboCliques per pivot) across the values 1, 2, and 3. The default $K_1$ values are set to 2000, 1000, and 1000 for 3DMatch+FPFH, 3DMatch+FCGF, and 3DLoMatch+Predator, respectively, while the default $K_2$ is consistently set to 2 across all three configurations.}

\PAR{Effect of Compatibility Threshold.} Rows 1-9 of \cref{table: Ablation_study} illustrate the effect of $\tau$ on registration performance. At $\tau = 0.0005$~m (row~1), RR decreases significantly (e.g., 66.30\% compared to 84.10\% for the default 3DMatch+FPFH configuration) due to overly restrictive thresholds that exclude compatibility between inliers. Performance peaks at approximately $\tau = 0.012$~m and declines beyond 0.2~m across three configurations. Notably, RR exhibits stability near the optimal $\tau$, as exemplified by the 3DMatch+FPFH configuration, where RR consistently exceeds 82\% when $\tau$ ranges from 0.01~m to 0.08~m.

\PAR{Efficiency Improvement with Directed Graphs.}  
Analysis of rows 10 vs. 23 in \cref{table: Ablation_study} demonstrates three key findings:  
(1) Directed graphs enhance RR for 3DMatch+FPFH (+0.49\%) and 3DLoMatch+Predator (+0.18\%). This improvement likely stems from their lower inlier ratios, where directed O2Graph more effectively eliminates redundant TurboCliques while preserving geometrically consistent ones.  
(2) Directed graphs significantly improve CPU efficiency, reducing 3DMatch+FPFH processing time by 67.88 ms. 
This acceleration occurs because directed edge constraints reduce the average number of node neighbors, 
thereby decreasing the required iterations for clique searches.
(3) GPU efficiency exhibits minimal variation, with a difference of only 0.69 ms on 3DMatch+FPFH. \red{This results from the near-$\mathcal{O}(1)$ \redll{parallel} time complexity of PGS, rendering it largely insensitive to graph structure variations.}

\PAR{Effect of Pivot Number ($K_1$).}  
Rows 11-16 of \cref{table: Ablation_study} demonstrate that RR improves with increasing $K_1$ until it stabilizes (e.g., convergence occurs when $K_1 \geq 250$ for 3DMatch+FCGF). Larger $K_1$ values increase the number of generated hypotheses, enhancing the likelihood of inlier detection. However, beyond this convergence threshold, the computational cost rises linearly with $\mathcal{O}(N)$ complexity, yielding minimal improvement in RR. Additionally, GPU performance slightly declines at higher $K_1$ due to limitations in data transfer efficiency.

We observe that TurboReg achieves a sufficiently high RR when $K_1 = 1000$. For instance, under the 3DMatch+FCGF and 3DLoMatch configurations, TurboReg attains the highest RR at $K_1 = 1000$, while the 3DMatch+FPFH configuration achieves the second-highest RR. Notably, in this setup, TurboReg generates only 2000 hypotheses ($K_1 \cdot K_2 = 2000$), which is significantly lower than the number of hypotheses required by RANSAC, often reaching millions.

\PAR{Effect of $K_2$.}
Rows 17-19 of \cref{table: Ablation_study} detail the effect of $K_2$ on TurboReg's performance. Experimental results indicate that TurboReg achieves its highest computational speed at $K_2 = 1$, with performance gradually decreasing as $K_2$ increases. At $K_2 = 3$, the computational time overhead peaks due to the generation of additional hypotheses, and the RR slightly declines, possibly due to introduced noise. Nevertheless, across all three datasets evaluated, optimal performance is consistently achieved when $K_2 = 2$.

\PAR{Summary of Parameter Settings.} To assist readers in better tuning the TurboReg parameters, we summarize the key findings from the ablation studies. The main parameters of TurboReg are  three: 
(1) Compatibility Threshold ($\tau$): This is the most critical parameter in TurboReg, as it directly influences the performance. 
Based on the definition of TurboClique and the ablation study on rows 1-9 of \cref{table: Ablation_study}, its value should be smaller than the resolution of the point clouds.
We recommend initializing $\tau$ at approximately $0.25 \times$ the point cloud resolution 
(e.g., the resolution of the 3DMatch dataset is 5 cm and $\tau$ is initially set to 1.25 cm and then fine-tuned to 1.2 cm).
(2) Pivot Number ($K_1$): This parameter primarily affects the time complexity of TurboReg. 
In theory, increasing the value of $K_1$ improves performance, but at the cost of slower speed.
Our experiments show that a value of $K_1 = 1000$ strikes a good balance between performance and speed. (3) TurboCliques Per Pivot ($K_2$): Similar to $K_1$, this parameter affects the number of hypotheses. We recommend setting the default value of $K_2 = 2$.

\section{Conclusion}
\label{sec:conclusion}

In this paper, we propose a novel approach to tackling the challenges of slow speed in point cloud registration while preserving high accuracy. We introduce a new type of clique, termed TurboClique, which is both lightweight and stable for transformation estimation. To ensure a high inlier ratio in the identified TurboCliques, we present the Pivot-Guided Search (PGS) algorithm. Owing to the lightweight design of TurboClique, our algorithm achieves real-time runtime performance. Experimental results demonstrate that our method attains state-of-the-art performance across several datasets while achieving the fastest registration speed.

\PAR{Acknowledgements.}
This work was supported in part by the National Natural Science Foundation of China (NSFC) under Grant 42271444 and Grant 42030102.

{\small
\bibliographystyle{ieeenat_fullname}
\bibliography{11_references}
}

\ifarxiv \clearpage \appendix \ifarxiv \maketitlesupplementary \fi

In this supplementary material, we first provide additional analyses to elaborate on concepts introduced in the main paper. Specifically: 
(1) \cref{suppsec: wtus} details why TurboClique employs a stringent compatibility threshold. (1.1) \cref{suppsec: pcis} elaborates on pairwise compatibility-induced stability. 
(1.2) \cref{suppsec: ev} and \cref{supsec: gi} illustrate the application of pairwise compatibility-induced stability in the design of TurboClique through experimental validation and geometric intuition, respectively.
(2) \cref{suppsec: niscs} offers a detailed numerical interpretation of the SC$^2$ scores.
(3) \cref{supsec: unique} proves the Unique Assignment Property of TurboClique in the O2Graph. 
(4) \cref{appsec:code} presents the Tensor-style pseudo-code for PGS.

Next, we introduce foundational concepts to enhance the completeness. (1) \cref{suppsec: clique} defines the concepts of clique and maximal clique. (2) \cref{suppsec: vdlse} provides the derivation of the variance of the Least Squares Estimator.

Finally, we present additional experiments. 
(1) \cref{exp:idea_rt}  provides a deep understanding experiment for searched TurboClique
(2) \cref{supsec: cptc} provides the runtime of TurboReg components. (3) \cref{suppsec: fca} analyzes the failure cases of TurboReg. (4) \cref{supsec: qv} visualizes qualitative examples not included in the main paper.

\section{Why TurboClique using Stringent Compatibility Threshold ?}
\label{suppsec: wtus}

In this section, we introduce the rationale behind the stringent compatibility threshold employed by TurboClique. We first elaborate on the concept of pairwise compatibility-induced stability in \cref{suppsec: pcis}. Furthermore, we demonstrate how this stability principle is incorporated into the design of TurboClique through experimental validation in \cref{suppsec: ev} and geometric intuition in \cref{supsec: gi}.

\subsection{Pairwise Compatibility-induced Stability}
\label{suppsec: pcis}

This section explains pairwise compatibility-induced stability by analyzing the relationship between matching noise variance and the spatial compatibility constraint. The analysis demonstrates that smaller $\tau$ values enhance pairwise compatibility-induced stability, enabling 3-cliques to achieve stability comparable to larger cliques (e.g., maximal cliques). We also provide experimental and intuitive analyses for a comprehensive understanding.

Given a matching set $\mathcal{M} = \{ \boldsymbol{m}_{i}\}_{i=1}^N$, where, $\boldsymbol{m}_{i}=(\boldsymbol{x}_i, \boldsymbol{y}_i)$ and $\boldsymbol{x}_i, \boldsymbol{y}_i \in \mathbb{R}^3$ represent source and target keypoints, respectively, the matching relationship is defined as $\boldsymbol{y}_i = \mathbf{T}(\boldsymbol{x}_i) + \boldsymbol{r}_i$. Here, $\mathbf{T}(\cdot)$ denotes the rigid transformation (including rotation and translation), and $\boldsymbol{r}_i$ represents noise in the matching process. We assume $\boldsymbol{r}_i \sim \mathcal{N}(\mathbf{0}, \sigma^2 \mathbf{I}_{3 \times 3})$, indicating that the noise follows an independent, zero-mean, isotropic Gaussian distribution with variance $\sigma^2$.

In the absence of constraints between matches, the distribution of $\boldsymbol{r}_i$ remains entirely random, with noise uniformly distributed across three-dimensional space. We then analyze the influence of introducing spatial compatibility constraints, defined as follows for any $\boldsymbol{m}_{i} \in \mathcal{M}$:
\begin{equation}
	\left| \| \boldsymbol{y}_i - \boldsymbol{y}_j \| - \| \boldsymbol{x}_i - \boldsymbol{x}_j \| \right| \leq \tau,
	\label{eq:transformed_constraint}
\end{equation}
where $\tau \geq 0$ represents the compatibility threshold, limiting the distance difference between matching pairs. This constraint reduces the randomness of $\boldsymbol{r}_i$, narrows the noise distribution, and yields an effective variance $\sigma_{\text{eff}}^2$ that is potentially smaller than the initial variance $\sigma^2$.

For a precise analysis, we define $\boldsymbol{d}_{ij} = \boldsymbol{x}_i - \boldsymbol{x}_j$ and $\boldsymbol{e}_{ij} = \boldsymbol{r}_i - \boldsymbol{r}_j$. 
Leveraging the distance-preserving property of rigid transformations, which ensures that $\|\mathbf{T}(\boldsymbol{x}_i) - \mathbf{T}(\boldsymbol{x}_j)\| = \|\boldsymbol{x}_i - \boldsymbol{x}_j\|$, we rewrite \cref{eq:transformed_constraint} as:
\begin{equation}
\left| \| \boldsymbol{d}_{ij} + \boldsymbol{e}_{ij} \| - \| \boldsymbol{d}_{ij} \| \right| \leq \tau.
\label{eq:cons2}
\end{equation}
Initially, $\boldsymbol{e}_{ij}$ follows a Gaussian distribution $\mathcal{N}(\mathbf{0}, 2\sigma^2 \mathbf{I}_{3 \times 3})$. However, the constraint in Eq.
\eqref{eq:cons2} 
limits the norm of
$\boldsymbol{e}_{ij}$, effectively truncating the joint distribution of $\boldsymbol{r}_i$ and $\boldsymbol{r}_j$. Consequently, the variance of this truncated distribution is smaller than that of the original, resulting in an effective variance $\sigma_{\text{eff}}^2 < \sigma^2$. Specifically, 
\cref{eq:cons2}
enforces
$\| \boldsymbol{d}_{ij} + \boldsymbol{e}_{ij} \|$ 
to lie within
the interval $[\| \boldsymbol{d}_{ij} \| - \tau, \| \boldsymbol{d}_{ij} \| + \tau]$. As $\tau$ decreases, this interval narrows, further restricting the possible values of $\boldsymbol{e}_{ij}$. In 
the limit
where $\tau \to 0$, the constraint reduces to $\| \boldsymbol{d}_{ij} + \boldsymbol{e}_{ij} \| = \| \boldsymbol{d}_{ij} \|$, which geometrically implies that $\boldsymbol{e}_{ij} \to \mathbf{0}$. This condition suggests that $\boldsymbol{r}_i \approx \boldsymbol{r}_j$, and given the zero-mean property of $\boldsymbol{r}_i$, it follows that $\boldsymbol{r}_i \to \mathbf{0}$. As a result, the noise distribution approaches a Dirac delta function, with $\sigma_{\text{eff}}^2 \to 0$. Thus, smaller values of $\tau$ progressively reduce $\sigma_{\text{eff}}^2$, ultimately approaching zero.

In summary, spatial compatibility reduces the randomness of $\boldsymbol{r}_i$, with its variance decreasing to zero as $\tau$ diminishes.

\begin{figure}[tp]
	\centering
	\includegraphics[width=\linewidth]{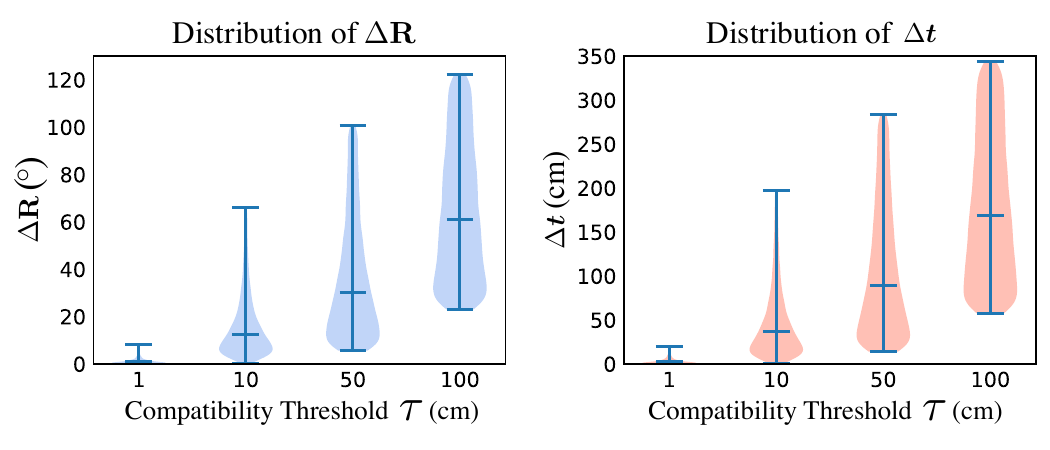}
	\caption{Distribution of discrepancies between transformations estimated from 3-clique and 10-clique configurations in terms of rotation and translation.}
	\label{fig:RRERTE}
	\vspace{-1.em}
\end{figure}

\begin{figure}[tp]
	\centering
	\includegraphics[width=\linewidth]{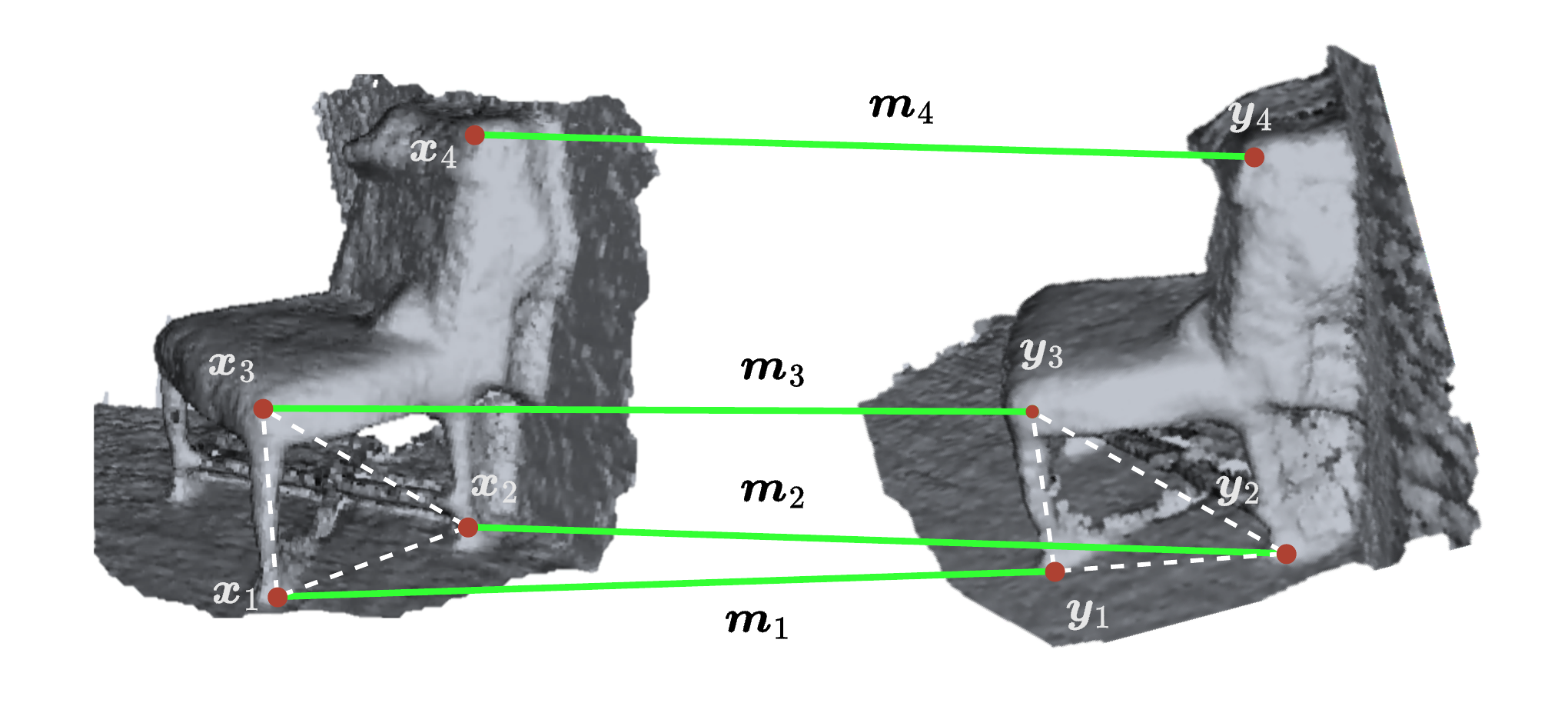}
	\caption{Demonstration of matches when $\tau=0$.}
	\label{fig:T34}
	\vspace{-1.em}
\end{figure}

\subsection{Experimental Validation}
\label{suppsec: ev}

The analysis above suggests that as $\tau$ approaches zero, pairwise compatibility-induced stability increases, compensating for the loss of data scaling stability in TurboClique due to a reduced number of matches. Consequently, rigid transformations derived from 3-cliques exhibit minimal differences compared to those from larger cliques, supporting TurboClique’s preference for a small $\tau$. This section empirically confirms this inference by demonstrating that transformation discrepancies between 3-cliques and 10-cliques decrease as $\tau$ diminishes. Specifically, we assess transformation discrepancies between estimates derived from 3-cliques and 10-cliques on the 3DMatch+FPFH dataset across increasing $\tau$ values. The procedure is outlined as follows:

\begin{enumerate}
	\item Set $\tau \in \{ 1\ \text{cm}, 10\ \text{cm}, 50\ \text{cm}, 100\ \text{cm} \}$ to construct compatibility graph $\mathbf{G}$;
	\item Extract all 10-cliques from $\mathbf{G}$ and compute multiple transformations $\mathbf{T}^{(10)} = (\mathbf{R}^{(10)}, \boldsymbol{t}^{(10)})$;
	\item For each 10-clique, estimate transformations $\{ \mathbf{T}^{(3)}_{k} \}_{k=1}^{K} = \{ (\mathbf{R}_{k}^{(3)}, \boldsymbol{t}_{k}^{(3)}) \}$ from all $K = \binom{10}{3}$ 3-clique subsets;
	\item Calculate rotation and translation errors:
	\begin{equation}
\Delta \mathbf{R}_{k} = \arccos\left(\frac{\text{tr}(\mathbf{R}^{(10)\top}\mathbf{R}^{(3)}_k) - 1}{2}\right),
\end{equation}
	\begin{equation}
\Delta \boldsymbol{t}_{k} = \|\boldsymbol{t}^{(10)} - \boldsymbol{t}^{(3)}_k\|_2;
\end{equation}
	\item Visualize error distributions across $\tau$ values in \cref{fig:RRERTE}.
\end{enumerate}

Results show negligible discrepancies at $\tau = 1$ cm ($< 0.1^\circ$ rotation, $< 0.5$ mm translation), with errors rising proportionally to $\tau$. This confirms that 3-cliques achieve accuracy comparable to larger cliques under tight thresholds, while significantly reducing computational complexity.

\subsection{Geometric Intuition}
\label{supsec: gi}

We further provide an intuitive analysis by examining the extreme case where $\tau = 0$. Three matches $\{ \boldsymbol{m}_{1}, \boldsymbol{m}_{2}, \boldsymbol{m}_{3} \}$, as shown in \cref{fig:T34}, form congruent triangles $\triangle \boldsymbol{x}_{1}\boldsymbol{x}_{2}\boldsymbol{x}_{3}$ and $\triangle \boldsymbol{y}_{1}\boldsymbol{y}_{2}\boldsymbol{y}_{3}$, uniquely determining the rigid transformation $\mathbf{T}^{(3)}$. Introducing a fourth match $\boldsymbol{m}_{4}$ that satisfies $\tau = 0$ compatibility with the initial trio results in a transformation $\mathbf{T}^{(4)}$ identical to $\mathbf{T}^{(3)}$, as $\boldsymbol{m}_{4}$ must conform to the existing geometric constraints. This principle applies to additional matches: any correspondence satisfying $\tau = 0$ preserves the original transformation. Thus, under ideal compatibility conditions (i.e., $\tau = 0$), 3-cliques fully encapsulate transformation information, rendering larger cliques unnecessary. In practice, however, a small, non-zero $\tau$ is adopted to account for sensor noise and matching imperfections, justifying our use of a modest compatibility threshold.

\section{Numerical Interpretation of SC² Scores}
\label{suppsec: niscs}

In Sec.~3.3 of the main paper, we claim that SC$^2$ scores quantify TurboClique density.
We now provide a brief explanation. The SC$^2$ score between matches $\boldsymbol{m}_i$ and $\boldsymbol{m}_j$ is defined as:
\begin{equation}
\hat{\mathbf{G}}_{ij} = \mathbf{G}_{ij} \sum_{k=1}^N \mathbf{G}_{ik} \cdot \mathbf{G}_{jk}, 
\end{equation}
where $\mathbf{G}_{ij} \in \{0, 1\}$ indicates spatial compatibility between $\boldsymbol{m}_i$ and $\boldsymbol{m}_j$. Two observations are as follows:
\begin{itemize}[leftmargin=2em, itemsep=0.5em, parsep=0.1em]
    \item If $\mathbf{G}_{ij} = 0$, then $\hat{\mathbf{G}}_{ij} = 0$, indicating no edge between $\boldsymbol{m}_{i}$ and $\boldsymbol{m}_{j}$. Consequently, the number of TurboCliques around $(\boldsymbol{m}_{i}, \boldsymbol{m}_{j})$ is zero.
    \item If $\mathbf{G}_{ij} = 1$, the summation counts nodes $k$ where $\mathbf{G}_{ik} = \mathbf{G}_{jk} = 1$. Each such $k$ forms a TurboClique $\{\boldsymbol{m}_i, \boldsymbol{m}_j, \boldsymbol{m}_k\}$, making $\hat{\mathbf{G}}_{ij}$ equal to the number of TurboCliques containing the edge $(i, j)$.
\end{itemize}

Combining these cases, the value of $\hat{\mathbf{G}}_{ij}$ represents the number of TurboCliques associated with $\boldsymbol{m}_{i}$ and $\boldsymbol{m}_{j}$.

\section{Unique Assignment Property of TurboClique}
\label{supsec: unique}

In this section, we demonstrate how the O2Graph eliminates the redundant detection of TurboCliques by proving that each TurboClique can be uniquely assigned to a single pivot. 

Given a TurboClique around $\pi_{z}$, denoeted as  $\text{TC}(\pi_{z}) = \{ \boldsymbol{m}_{z_{1}}, \boldsymbol{m}_{z_{2}}, \boldsymbol{m}_{z_{3}} \}$, where $z_{1} < z_{2} < z_{3}$ (without loss of generality), the O2Graph defines edge directions from lower-indexed to higher-indexed nodes. This implies:
\begin{itemize}[leftmargin=2em, itemsep=0.5em, parsep=0.1em]
	\item $\mathcal{N}(\boldsymbol{m}_{z_{1}}) = \{ \boldsymbol{m}_{z_{2}}, \boldsymbol{m}_{z_{3}} \}$,
	\item $\mathcal{N}(\boldsymbol{m}_{z_{2}}) = \{ \boldsymbol{m}_{z_{3}} \}$,
	\item $\mathcal{N}(\boldsymbol{m}_{z_{3}}) = \emptyset$, 
\end{itemize}
where $\mathcal{N}(\cdot)$ denotes the set of neighboring nodes in the compatibility graph.

Next, we analyze three possible pivot cases to show that only one case detects $\text{TC}(\pi_{z})$:
\begin{itemize}[leftmargin=2em, itemsep=0.5em, parsep=0.1em]
{
\item \textbf{Case 1: $\pi_{z} = (\boldsymbol{m}_{z_{2}}, \boldsymbol{m}_{z_{3}})$}: Since $\boldsymbol{m}_{z_{1}} \notin \mathcal{N}(\boldsymbol{m}_{z_{2}})$ and $\boldsymbol{m}_{z_{1}} \notin \mathcal{N}(\boldsymbol{m}_{z_{3}})$, this pivot cannot detect $\text{TC}(\pi_{z})$.
\item \textbf{Case 2: $\pi_{z} = (\boldsymbol{m}_{z_{1}}, \boldsymbol{m}_{z_{3}})$}: Since $\boldsymbol{m}_{z_{2}} \notin \mathcal{N}(\boldsymbol{m}_{z_{3}})$, this pivot cannot form $\text{TC}(\pi_{z})$ with $\boldsymbol{m}_{z_{2}}$.
\item  \textbf{Case 3: $\pi_{z} = (\boldsymbol{m}_{z_{1}}, \boldsymbol{m}_{z_{2}})$}: Here, $\boldsymbol{m}_{z_{3}} \in \mathcal{N}(\boldsymbol{m}_{z_{1}}) \cap \mathcal{N}(\boldsymbol{m}_{z_{2}})$, enabling the formation of $\text{TC}(\pi_{z})$.
}
\end{itemize}

Since any three matches can form at most the three above pivot configurations, and only the pivot consisting of the two lowest-indexed nodes detects a TurboClique, this proves that each TurboClique is uniquely assigned to a single pivot.

\section{Tensor-style Pseudo-code of PGS}
\label{appsec:code}

We present the Tensor-style pseudo-code of the PGS algorithm in Algorithm~\ref{algo:pgs-gpu}.

\begin{algorithm}[t]
	\SetAlgoLined	
	\textbf{Input: } Weighted graph: $\bar{\mathbf{G}} \in \mathbb{R}^{N \times N}$; number of pivots $K_1 \in \mathbb{N}^+$; number of TurboCliques for each pivot $K_2 \in \mathbb{N}^+$  \\
	\textbf{Output: } TurboClique set $\mathbf{C} \in \{1, \dots, N\}^{K_1K_2 \times 3}$  \\
	\annot{Select top-$K_1$ edges as pivots} \\
	$\mathbf{P} \gets \text{TopKEdges}(\bar{\mathbf{G}}, K_1)\quad$ \\
	\annot{Common neighbors (mask) for each pivot} \\
	$\mathbf{M} \gets (\bar{\mathbf{G}}[\mathbf{P}[:, 0]] > 0) \odot (\bar{\mathbf{G}}[\mathbf{P}[:, 1]] > 0)$ \\
	\annot{TurboClique weights for each TurboClique} \\
	$\mathbf{S} \gets \bar{\mathbf{G}}[\mathbf{P}[:, 0], \mathbf{P}[:, 1]] + (\bar{\mathbf{G}}[\mathbf{P}[:, 0]] + \bar{\mathbf{G}}[\mathbf{P}[:, 1]])$ \\
	$\mathbf{S}' \gets \mathbf{S} \odot \mathbf{M}$\\
	\annot{Top-$K_2$ TurboCliques for each pivot} \\
	$\mathbf{Z} \gets \text{ColumnTopK}(\mathbf{S}', K_2)$ 	\\
	\annot{Assemble TurboCliques: (pivots, third matches)} \\
	$\mathbf{C} \gets \text{zeros}(K_1 \cdot K_2, 3)$ \\
	\For{$i \gets 0 \ \text{to} \ K_2$} {
		\annot{Assign first two matches} \\
		$\mathbf{C}[(i \times K_1):((i + 1) \times K_1), :2] \gets \mathbf{P}$ \\
		\annot{Assign third match} \\
		$\mathbf{C}[(i \times K_1):((i + 1) \times K_1), 2] \gets \mathbf{Z}$
	}
	\Return $\mathbf{C}$
	\caption{Pivot-Guided Search Algorithm (Tensor-style)}
    \label{algo:pgs-gpu}
\end{algorithm}

\section{Supporting Theorems and Derivations}
\label{suppsec:std}
To ensure the completeness of this paper, this section provides foundational theorems and derivations that support the main analysis.

\subsection{Definition of Clique and Maximal Clique}
\label{suppsec: clique}

\begin{figure}[tp]
	\centering
	\includegraphics[width=0.45\linewidth]{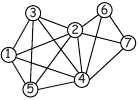}
	\caption{Undirected graph for demonstration.}
	\label{suppfig: ungraph}
	\vspace{-1.em}
\end{figure}

Figure \cref{suppfig: ungraph} depicts a graph with 7 vertices, denoted as $\mathcal{G}$. A clique is a complete subgraph $\mathcal{C} \subseteq \mathcal{G}$ where every pair of distinct vertices is adjacent:
\begin{equation}
\forall \boldsymbol{u}, \boldsymbol{v} \in \mathcal{C}, (\boldsymbol{u}, \boldsymbol{v}) \in E(\mathcal{G}),
\end{equation}
where $E(\mathcal{G})$ represents the edge set of $\mathcal{G}$. For example, the vertices $\{1, 3, 5\}$ are fully connected, forming a 3-clique. Similarly, $\{2, 4, 6, 7\}$ constitutes a 4-clique.
For example, the vertices $\{ 1, 3, 5 \}$ are fully connected, forming a 3-clique. Similarly, $\{ 2, 4, 6, 7 \}$ forms a 4-clique.

The maximal clique is defined as a clique that cannot be extended by including any adjacent vertex:
\begin{equation}
\nexists \boldsymbol{w} \in \mathcal{G} \setminus \mathcal{C} \text{ such that } \mathcal{C} \cup \{\boldsymbol{w}\} \text{ forms a clique}.
\end{equation}
For instance, the 5-clique $\{1, 2, 3, 4, 5\}$ is maximal because the remaining vertices $\{6, 7\}$ cannot be added to form a larger clique. Similarly, $\{2, 4, 6, 7\}$ is a maximal 4-clique.

\subsection{Variance of LS Estimator}
\label{suppsec: vdlse}

This section derives the variance of the least squares (LS) estimator in a standard linear regression framework. Consider the linear regression model:
\begin{equation}
Y = X\beta + \epsilon,
\end{equation}
where $Y$ is the response variable, $X$ is the design matrix, $\beta$ is the coefficient vector, and $\epsilon$ is the error term. We assume the errors satisfy:
\begin{equation}
\mathbb{E}[\epsilon | X] = 0, \quad \text{Var}(\epsilon | X) = \sigma^2 I_n.
\end{equation}
The ordinary least squares (OLS) estimator for $\beta$ is:
\begin{equation}
\hat{\beta} = (X'X)^{-1} X' Y.
\end{equation}
The variance of $\hat{\beta}$ is computed as:
\begin{equation}
\text{Var}(\hat{\beta} | X) = \text{Var} \left( (X'X)^{-1} X' Y \mid X \right).
\end{equation}
Substituting $Y = X\beta + \epsilon$ and applying variance properties:
\begin{equation}
\text{Var}(\hat{\beta} | X) = (X'X)^{-1} X' \text{Var}(\epsilon | X) X (X'X)^{-1}.
\end{equation}
Given $\text{Var}(\epsilon | X) = \sigma^2 I_n$, this simplifies to:
\begin{equation}
\text{Var}(\hat{\beta} | X) = \sigma^2 (X'X)^{-1}.
\end{equation}
This result indicates that the variance of the LS estimator depends on the noise variance $\sigma^2$ and the design matrix $X$. Notably, a smaller $\sigma^2$ or a larger sample size (reflected in $X$) reduces the variance.

\section{More Experiments}
\label{exp:more_exp}

\subsection{Understanding the Searched TurboCliques }
\label{exp:idea_rt}

\begin{table}[t]
	\centering
	\renewcommand{\arraystretch}{1.1}
	\renewcommand\tabcolsep{4pt}
	\resizebox{1.0\columnwidth}{!}{
		\begin{tabular}{cc|c|c|c|cccc}
			\hline
			                                      & \multirow{2}{*}{Metrics} & \multirow{2}{*}{RR (\%)} & \multirow{2}{*}{TQRR (\%)} & \multirow{2}{*}{ICRR (\%)} & \multicolumn{4}{c}{TKRR (\%)}                         \\
			                                      &                          &                     &                       &                       & @2                       & @3    & @5    & @50   \\
			\hline
			\multirow{3}{*}{\rotatebox{90}{FPFH}} & IN                       & 84.10               & 93.72                 & 70.48                 & 85.97                    & 86.50 & 87.77 & 92.38 \\
			                                      & MSE                      & 82.99               & 99.94                 & 68.90                 & 83.77                    & 84.43 & 85.34 & 90.04 \\
			                                      & MAE                      & 83.43               & 99.94                 & 69.26                 & 84.22                    & 84.87 & 85.79 & 90.51 \\
			\hline
			\multirow{3}{*}{\rotatebox{90}{FCGF}} & IN                       & 93.59               & 97.66                 & 90.24                 & 94.06                    & 94.45 & 94.91 & 96.66 \\
			                                      & MSE                      & 93.47               & 99.94                 & 89.73                 & 93.73                    & 93.86 & 94.24 & 95.86 \\
			                                      & MAE                      & 93.35               & 99.94                 & 89.84                 & 93.85                    & 93.98 & 94.37 & 95.98 \\
			\hline
		\end{tabular}}
		\caption{\red{Ranking-based Registration Recall Evaluation on the 3DMatch Dataset.} (1) TQRR evaluates whether the best transformation outperforms the ground truth transformation under the corresponding metrics. (2) ICRR assesses whether the best TurboClique hypothesis consists of three inliers.  
		(3) TKRR determines whether the top-K hypotheses include a successfully registered rigid transformation.}
		\label{tab:ideal_reg}
\end{table}

To better understand TurboReg, we propose three ranking-based registration recall metrics to analyze the $K_1K_2$ TurboCliques identified by PGS.
Specifically, we first rank the $K_1K_2$ transformation hypotheses based on inlier number (IN), mean absolute error (MAE), and mean squared error (MSE). The three metrics are defined as follows:  (1) Transformation Quality Registration Recall (TQRR): The proportion of cases where the top-1 hypothesis achieves a score equal to or exceeding the ground-truth transformation.  Inlier-Clique Registration Recall (ICRR): The proportion of cases where the top-1 hypothesis clique contains only inliers. (2) Top-$K$ Hypothesis Registration Recall (TKRR): The proportion of cases where at least one valid transformation exists among the top-$K$ hypotheses.   Results are summarized in Table \ref{tab:ideal_reg}.

\PAR{Discussion of TQRR.} From \cref{tab:ideal_reg}, TQRR consistently exceeds RR by significant margins. For instance, when ranked by IN, TQRR surpasses RR by 9.62\%, indicating that erroneous rigid transformations with higher consistency scores than the ground truth are frequently selected during model estimation. This suggests that the ground-truth transformation does not always align with the maximum consistency assumption, potentially due to:  
(1) Sampling Error: Discrete keypoint sampling or insufficient sampling density causing deviations between the ground truth and maximum consistency transformations.  
(2) Scene Ambiguity: Repetitive structures (e.g., identical objects) leading to ambiguous alignments.  

Notably, TurboReg achieves 99.94\% (1622/1623) TQRR under MAE and MSE metrics, demonstrating its ability to prioritize highly consistent hypotheses over the ground truth in nearly all cases.

\PAR{Discussion of ICRR.}  
\cref{tab:ideal_reg} reveals that ICRR is consistently lower than RR, indicating that many cliques containing outliers still produce successful registrations. These findings demonstrate that even cliques with outliers can yield correct registrations.

\PAR{Discussion of TKRR.}  
As $K$ increases, TKRR improves significantly and eventually surpasses RR (Table \ref{tab:ideal_reg}). This indicates that the correct transformation is more likely to reside among the top-$K$ transformations rather than exclusively in the top-1. However, conventional methods typically select the top-1 transformation based on ranking, implying that our model selection strategy may impose performance limitations. This reliance on top-1 selection often overlooks potentially correct transformations within the broader top-$K$ set, highlighting a key bottleneck in registration.

\PAR{Summary.}   In summary, we demonstrate that TurboReg excels at identifying TurboCliques with a high inlier ratio, characterized by high IN and lower MSE/MAE. However, the correct transformation may not always be selected due to the inherent limitation of choosing only the top candidate, which constrains the overall performance of the registration algorithm despite its ability to generate high-scoring cliques.

\begin{table}[t]
	\centering
	\renewcommand{\arraystretch}{1.1}
	\renewcommand\tabcolsep{1pt}
	\resizebox{1.0\columnwidth}{!}{%
		\begin{tabular}{c|c|ccc|c}
			\hline
			Device               & Methods    & O2Graph Construction & PGS             & Model Estimation & Total  \\
			\hline
			\multirow{2}{*}{CPU}   & Ours (0.5) & 276.05 (88.33\%) & 30.33 (9.70\%) & 6.13 (1.96\%) & 312.50 \\
									& Ours (2K) & 277.26 (67.02\%) & 73.26 (17.71\%) & 63.17 (15.27\%) & 413.68 \\

			\hline
			\multirow{2}{*}{GPU} & Ours (0.5) & 0.04 (0.25\%)        & 11.36 (71.71\%) & 4.44 (28.05\%)   & 15.84  \\
			                     & Ours (2K)  & 0.05 (0.25\%)        & 11.88 (60.80\%) & 7.61 (38.95\%)   & 19.54  \\

			\hline
		\end{tabular}}
	\caption{Average consumed time (ms) per point cloud pair on the 3DMatch+FPFH dataset across CPU and GPU implementations.}
	\label{supptab: time_cpu}%
\end{table}%

\subsection{Comparison with MAC hypotheses.}
Following \cite{zhang20233d}, we evaluate the quality of the generated hypotheses by comparing those produced by MAC and TurboReg against the ground truth transformation. The results, shown in \cref{tab:truehypo}, indicate that under the same number of hypotheses, our method yields a higher proportion of correct hypotheses.

\begin{table}[t]
	\centering
	\resizebox{\linewidth}{!}{
		\begin{tabular}{c|cc|cc|cc|cc}
			\hline
			\multirow{3}{*}{} & \multicolumn{4}{c|}{3DMatch} & \multicolumn{4}{c}{3DLoMatch}                                                                                                            \\ \cline{2-9}
			\#hypotheses     & \multicolumn{2}{c|}{FPFH}    & \multicolumn{2}{c|}{FCGF}     & \multicolumn{2}{c|}{FPFH} & \multicolumn{2}{c}{FCGF}                                                     \\
			%
			                  & 3DMAC                          & Ours                          & 3DMAC                       & Ours                     & 3DMAC   & Ours           & 3DMAC    & Ours            \\
			\hline
			100               & 50.67                        & \textbf{78.39}                & 61.92                     & \textbf{90.67}           & 12.22 & \textbf{23.19} & 30.47  & \textbf{52.11}  \\
			200               & 89.27                        & \textbf{151.12}               & 119.20                    & \textbf{178.99}          & 17.59 & \textbf{37.34} & 55.57  & \textbf{97.87}  \\
			500               & 162.41                       & \textbf{346.03}               & 269.06                    & \textbf{429.54}          & 23.32 & \textbf{45.43} & 109.32 & \textbf{206.49} \\
			1000              & 217.32                       & \textbf{598.01}               & 456.18                    & \textbf{777.29}          & 26.02 & \textbf{63.33} & 156.11 & \textbf{316.24} \\
			2000              & 254.13                       & \textbf{770.39}               & 669.32                    & \textbf{1034.39}         & 29.31 & \textbf{78.34} & 202.12 & \textbf{362.05} \\

			\hline
		\end{tabular}}
	\caption{Comparison of correct hypothesis counts}
	\label{tab:truehypo}
\end{table}

\subsection{Runtime of TurboReg Components}
\label{supsec: cptc}

This experiment investigates the temporal characteristics of TurboReg modules on the 3DMatch+FPFH dataset. Average execution times (ms) for CPU and GPU implementations are presented in \cref{supptab: time_cpu}.

Significant differences exist between CPU and GPU implementations. Focusing first on the CPU variant, the O2Graph Construction module dominates the processing time under both 0.5K and 2K pivot configurations. The PGS and Model Estimation modules exhibit a positive correlation between $K_{1}$ and runtime, since an increase in $K_{1}$ leads to a higher number of TurboCliques.

In GPU implementations, the O2Graph Construction time decreases drastically (e.g., merely 0.25\% of total runtime) due to parallel computation capabilities. 
Furthermore, the parallelized TurboClique search enables the PGS module to maintain near-constant execution time, resulting in approximately 12 ms for both $K_{1}=500$ and $2000$.
Conversely, the Model Estimation module demonstrates a linear scaling trend with $K_{1}$, as additional TurboCliques necessitate incremental transformation estimations.  

\begin{figure*}[hbt!]
	\centering
	\includegraphics[width=0.95\linewidth]{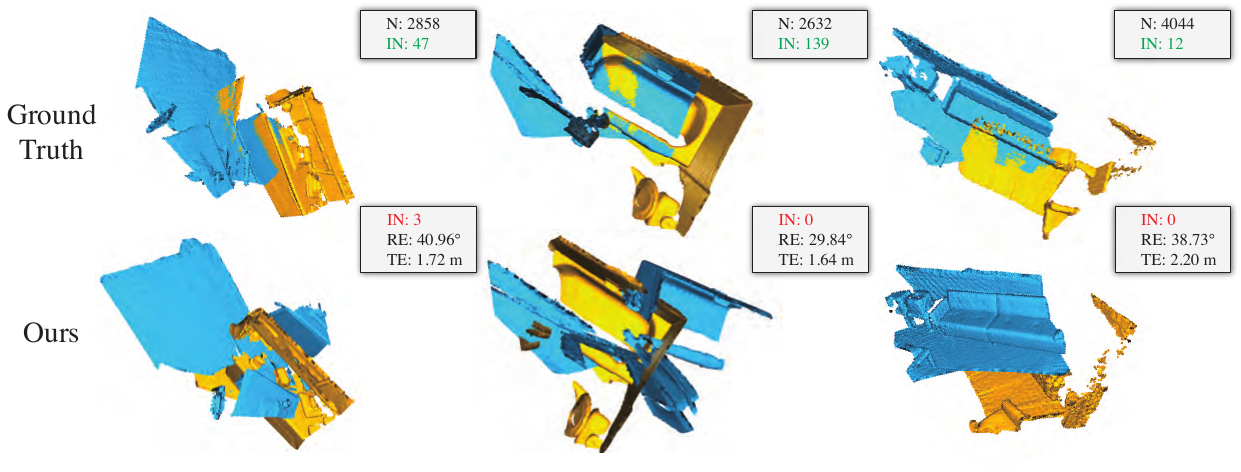}
	\caption{\textbf{Insufficient Consensus Correspondences.} \textcolor{red}{Red} indicates lower IN values, while \textcolor{green}{green} denotes higher IN values.}
	\label{fig: suppfig_failed_L1}
\end{figure*}

\begin{figure*}[hbt!]
	\centering
	\includegraphics[width=0.95\linewidth]{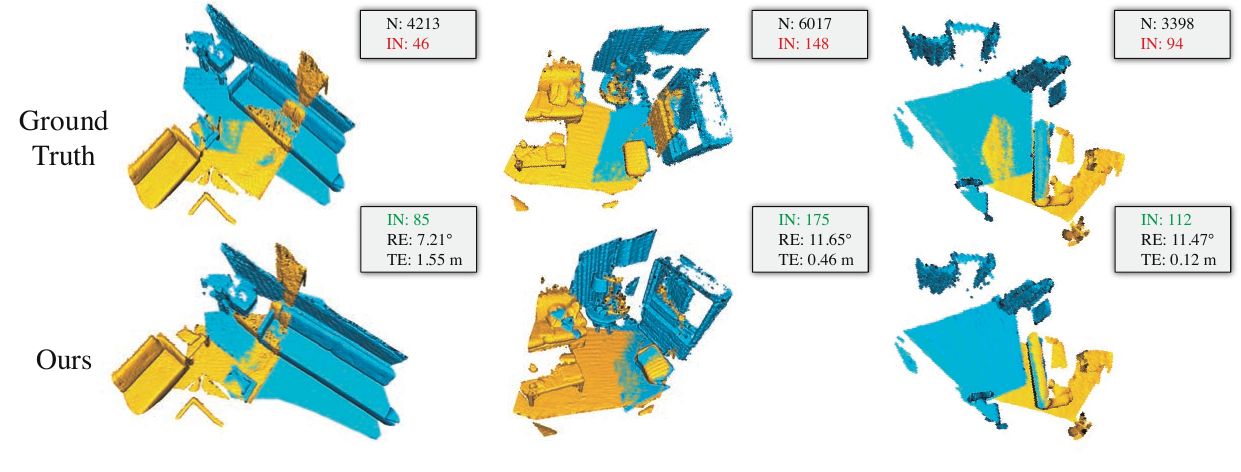}
	\caption{\textbf{Larger Consensus Set with Small Errors.} \textcolor{red}{Red} indicates lower IN values, while \textcolor{green}{green} denotes higher IN values.}
	\label{fig: suppfig_failed_L2}
\end{figure*}

\begin{figure*}[hbt!]
	\centering
	\includegraphics[width=0.95\linewidth]{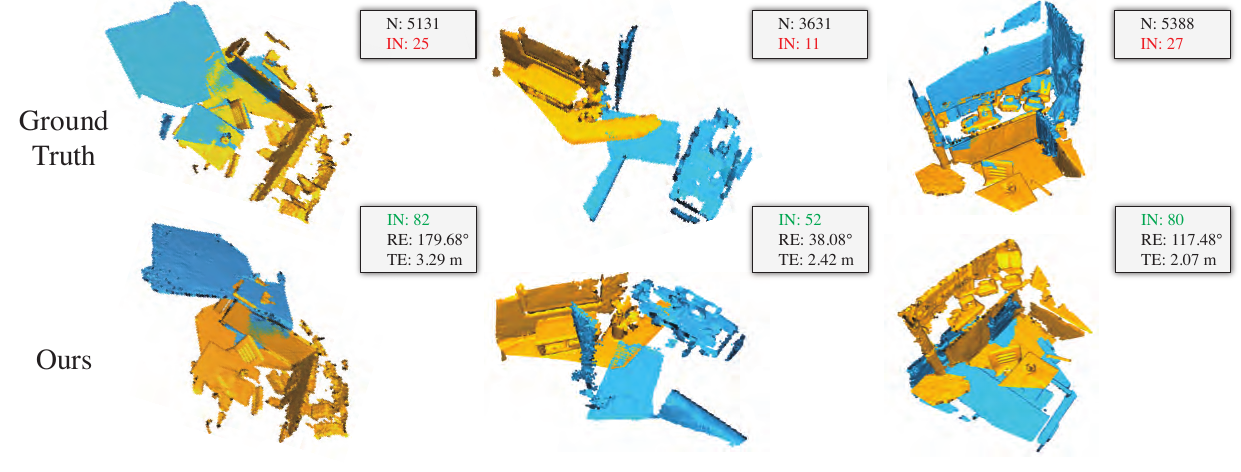}
	\caption{\textbf{Larger Consensus Set with Large Errors.} \textcolor{red}{Red} indicates lower IN values, while \textcolor{green}{green} denotes higher IN values.}
	\label{fig: suppfig_failed_L3}
\end{figure*}

\subsection{Failure Case Analysis}
\label{suppsec: fca}

In this section, we analyze the failure cases of TurboReg. We first review the definition of successful registration: registration is successful if the error between the estimated rigid transformation and the ground truth rigid transformation falls below specific thresholds. For 3DMatch and 3DLoMatch, the requirements are RE $\le 15^\circ$ and TE $\le 30$~cm. For the KITTI dataset, the requirements are RE $\le 5^\circ$ and TE $\le 60$~cm.

Next, we note that the estimated rigid transformation is selected based on the inlier number (IN), under the assumption that the correct rigid transformation corresponds to the maximum consensus set.

We classify instances that do not meet the successful registration criteria into three categories:
\begin{enumerate}
	\item \textbf{Insufficient Consensus Correspondences}: TurboReg fails to identify a sufficiently large set of consensus correspondences. This occurs in scenarios with extremely low overlap or strong symmetry, as illustrated in \cref{fig: suppfig_failed_L1}.
	\item \textbf{Larger Consensus Set but Incorrect Transformation}: The algorithm identifies a larger IN than that of the ground truth transformation, yet the result remains incorrect. This contradicts the maximum consensus set assumption. We categorize this scenario into two subcategories:
	\begin{enumerate}
		\item \textbf{Small Errors}: The estimated rigid transformation closely approximates the true rigid transformation, suggesting that registration is feasible, albeit with slightly larger errors. Due to the limited number of correct matches, the result is sensitive to noise, as illustrated in \cref{fig: suppfig_failed_L2}.
		\item \textbf{Large Errors}: The algorithm identifies a rigid transformation with a larger inlier set that still aligns visually, as shown in \cref{fig: suppfig_failed_L3}. This may occur because the matching pairs conform to an underlying geometric structure.
	\end{enumerate}
\end{enumerate}

\subsection{Qualitative Visualizations}
\label{supsec: qv}

Figs.~\ref{fig: suppfig_quan_3dmatch}-\ref{fig: suppfig_quan_kitti} illustrate qualitative visualizations of challenging registration pairs. 3DMAC and SC$^2$-PCR fail to achieve registration, whereas TurboReg successfully completes the registration task.

\begin{figure*}[hbt!]
	\centering
	\includegraphics[width=1.0\linewidth]{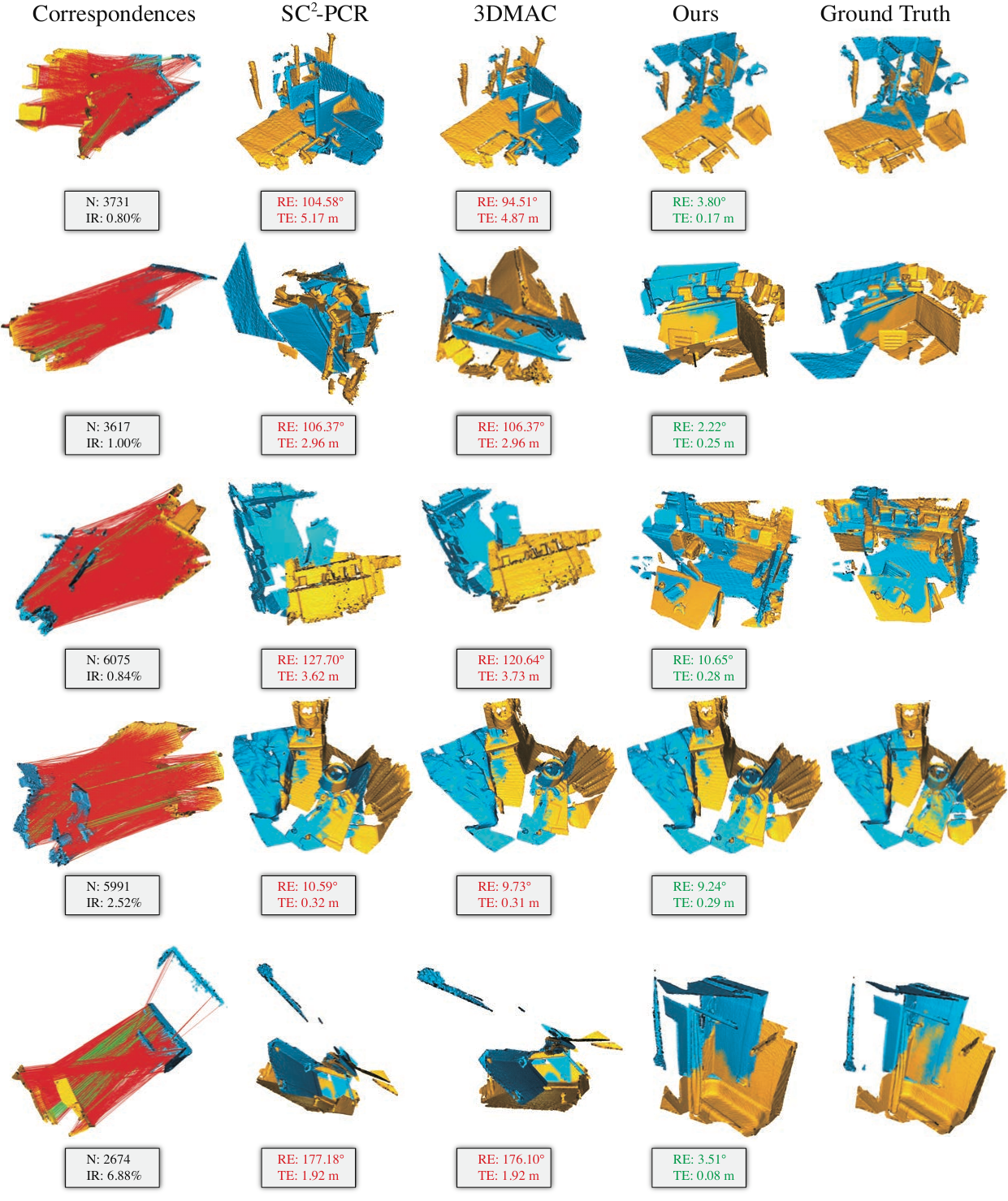}
	\caption{\textbf{Qualitative Comparison on 3DMatch.} \textcolor{red}{Red} and \textcolor{green}{green} represent failed and successful registrations, respectively.}
	\label{fig: suppfig_quan_3dmatch}
\end{figure*}

\begin{figure*}[hbt!]
	\centering
	\includegraphics[width=1.0\linewidth]{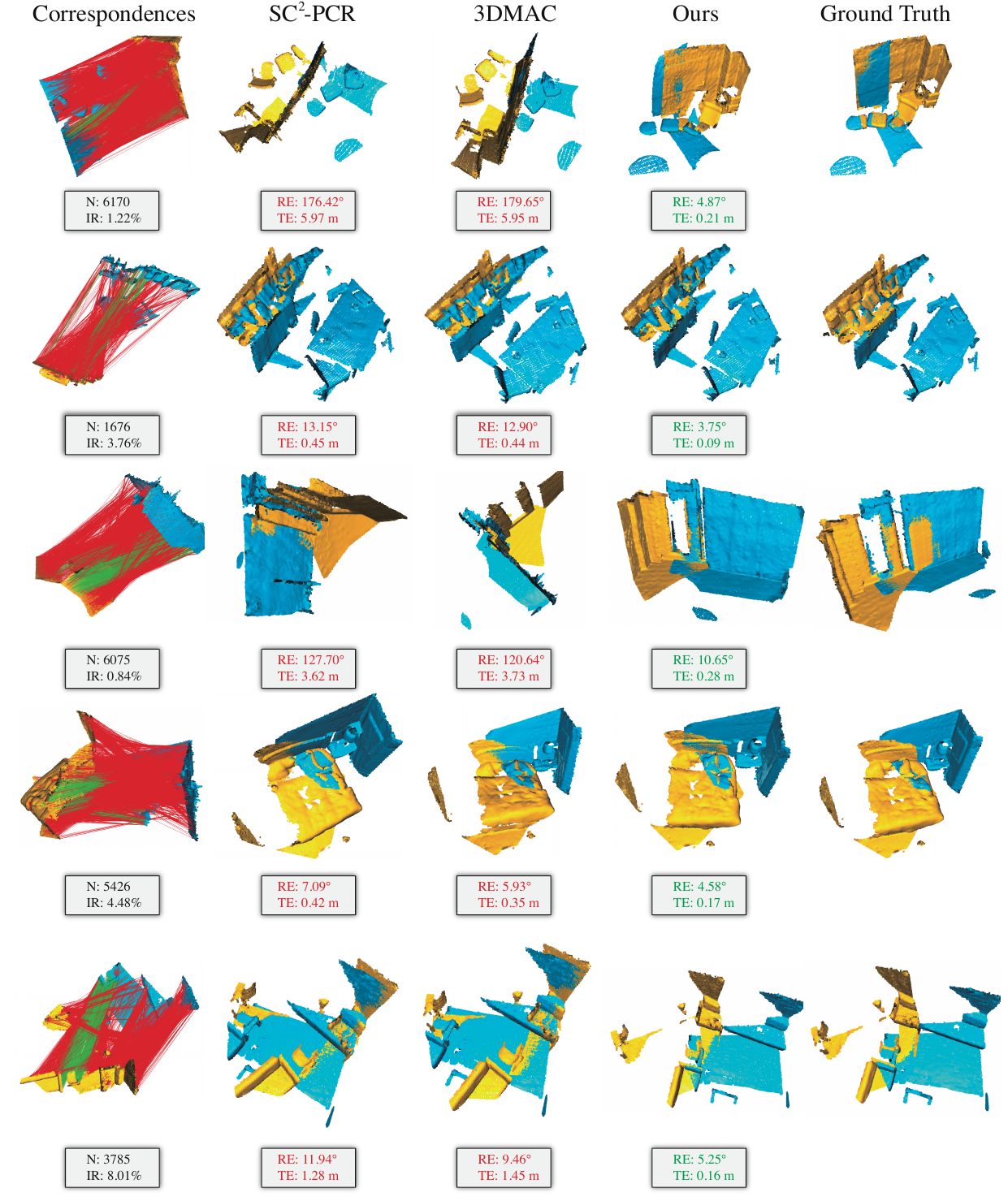}
	\caption{\textbf{Qualitative Comparison on 3DLoMatch.} \textcolor{red}{Red} and \textcolor{green}{green} represent failed and successful registrations, respectively.}
	\label{fig: suppfig_quan_3dlomatch}
\end{figure*}

\begin{figure*}[hbt!]
	\centering
	\includegraphics[width=1.0\linewidth]{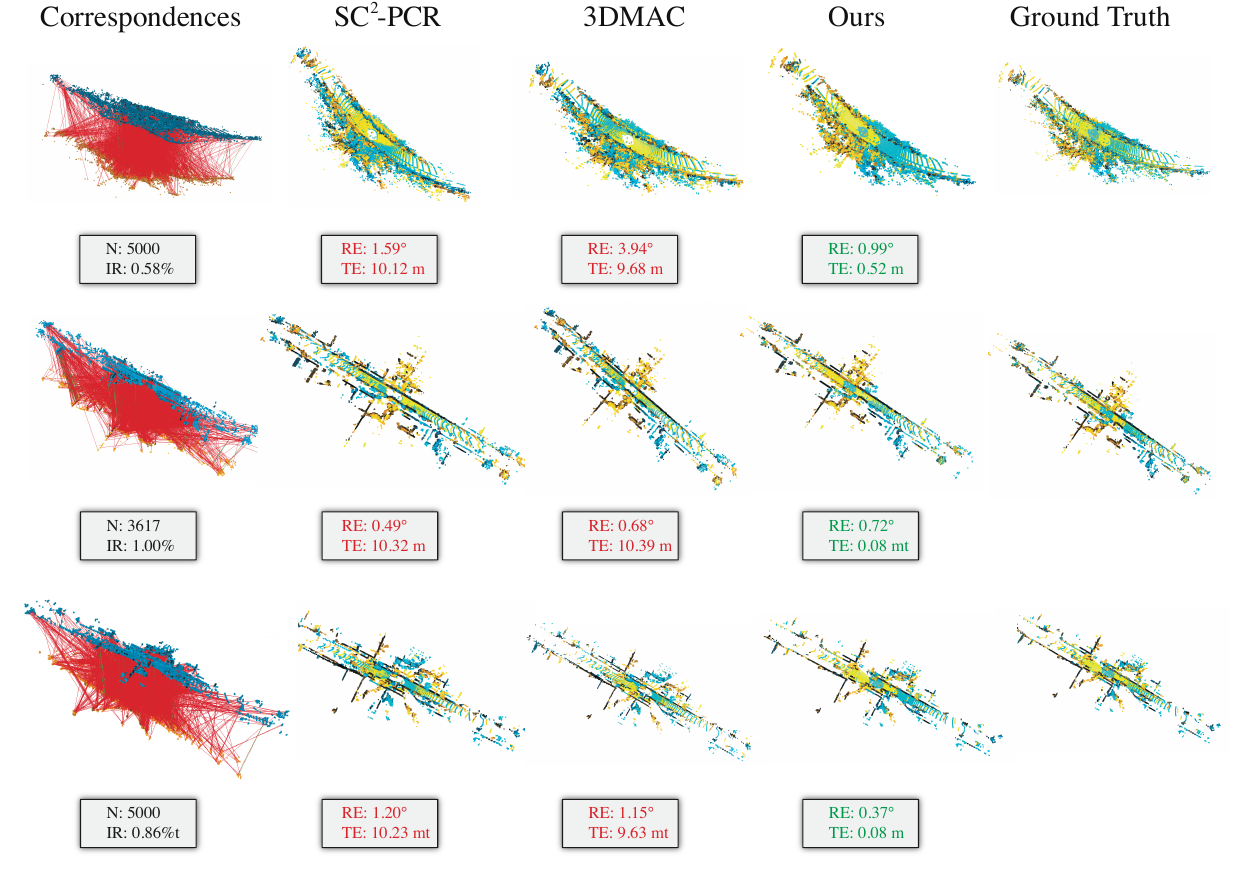}
	\caption{\textbf{Qualitative comparison on KITTI.} \textcolor{red}{Red} and \textcolor{green}{green} represent failed and successful registrations, respectively.}
	\label{fig: suppfig_quan_kitti}
\end{figure*}

 \fi

\end{document}